\documentclass[runningheads]{llncs}
\pdfoutput=1
 
\usepackage{eccv}

\usepackage{eccvabbrv}

\usepackage{booktabs}
\usepackage{graphicx}

\usepackage[accsupp]{axessibility}  

\usepackage{hyperref}

\usepackage{orcidlink}

\usepackage{soul}
\usepackage{amsmath}
\usepackage{wrapfig}
\usepackage{booktabs} 

\usepackage{amsmath,amsfonts,bm}

\def\eqref#1{equation~\ref{#1}}

\def\1{\bm{1}}

\def\rvx{{\mathbf{x}}}

\def\vmu{{\bm{\mu}}}

\def\valpha{{\bm{\alpha}}}
\def\vbeta{{\bm{\beta}}}

\def\vpsi{{\bm{\psi}}}
\def\vgamma{{\bm{\gamma}}}

\def\vc{{\bm{c}}}

\def\vq{{\bm{q}}}

\def\vs{{\bm{s}}}
\def\vt{{\bm{t}}}

\def\mR{{\bm{R}}}
\def\mS{{\bm{S}}}

\def\mSigma{{\bm{\Sigma}}}

\DeclareMathAlphabet{\mathsfit}{\encodingdefault}{\sfdefault}{m}{sl}
\SetMathAlphabet{\mathsfit}{bold}{\encodingdefault}{\sfdefault}{bx}{n}

\usepackage[misc]{ifsym}

\begin{document}

\title{HeadStudio: Text to Animatable Head Avatars with 3D Gaussian Splatting} 

\titlerunning{HeadStudio}


\author{
Zhenglin Zhou\inst{1, 2} \and
Fan Ma\inst{2} \and
Hehe Fan\inst{2} \and
Zongxin Yang\inst{2} \and
Yi Yang\inst{1, 2}$^{\dagger}$
} 

\authorrunning{Z. Zhou et al.}

\institute{
State Key Laboratory of Brain-machine Intelligence, Zhejiang University, China \and
ReLER, CCAI, Zhejiang University, China \\
\email{\{zhenglinzhou, mafan, hehefan, yangzongxin, yangyics\}@zju.edu.cn}\\
}

\maketitle
\def\thefootnote{$\dagger$}\footnotetext{Corresponding author.}


\begin{abstract}

Creating digital avatars from textual prompts has long been a desirable yet challenging task. 
Despite the promising results achieved with 2D diffusion priors, current methods struggle to create high-quality and consistent animated avatars efficiently.
Previous animatable head models like FLAME have difficulty in accurately representing detailed texture and geometry. Additionally, high-quality 3D static representations face challenges in semantically driving with dynamic priors.
In this paper, we introduce \textbf{HeadStudio}, a novel framework that utilizes 3D Gaussian splatting to generate realistic and animatable avatars from text prompts.
Firstly, we associate 3D Gaussians with animatable head prior model, facilitating semantic animation on high-quality 3D representations.
To ensure consistent animation, we further enhance the optimization from initialization, distillation, and regularization to jointly learn the shape, texture, and animation.
Extensive experiments demonstrate the efficacy of HeadStudio in generating animatable avatars from textual prompts, exhibiting appealing appearances. 
The avatars are capable of rendering high-quality real-time ($\geq 40$ fps) novel views at a resolution of 1024. 
Moreover, These avatars can be smoothly driven by real-world speech and video.
We hope that HeadStudio can enhance digital avatar creation and gain popularity in the community.
Code is at: \url{https://github.com/ZhenglinZhou/HeadStudio}.

\end{abstract}
\section{Introduction}
\label{sec:intro}

\begin{figure}[t]
    \centering
    \includegraphics[width=1.0\textwidth]{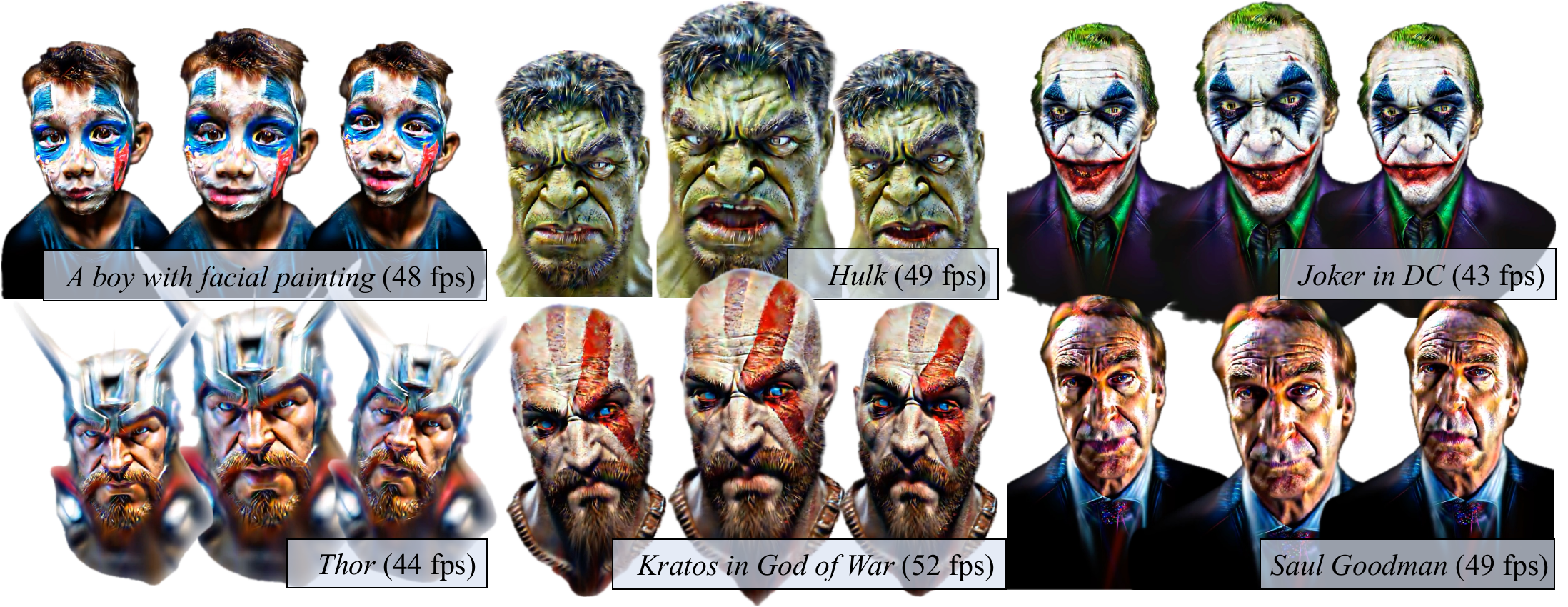}
    \caption{
    Text-based animatable avatars generation by \textbf{HeadStudio}. 
  With only one end-to-end training stage of 2 hours on 1 NVIDIA A6000 GPU, HeadStudio is able to generate animatable, high-fidelity and real-time rendering ($\geq 40$ fps) head avatars using text inputs.
    }
    \label{fig:teaser}
\end{figure}


With the development of deep learning, head avatar generation has improved significantly in recent years.
At first, the image-based methods~\cite{Chan2022eg3d,INSTA:CVPR2023} are proposed to reconstruct the photo-realistic head avatar of a person, given one or more views.
Recently, generative models (\eg~diffusion~\cite{stable_diffusion, zhang2023controlnet}) have made unprecedented advancements in high-quality text-to-image synthesis.
As a result, the research focus has been on text-based head avatar generation methods~\cite{han2023headsculpt, liu2023headartist}, which have shown superiority over image-based methods in convenience and generalization.

However, current text-based methods cannot combine high-quality and animation effectively. 
For instance, HeadSculpt~\cite{han2023headsculpt} leverages DMTet~\cite{shen2021deep} for high-quality optimization and creates highly detailed head avatars but is unable to be animated. 
TADA~\cite{liao2023tada} employs SMPL-X~\cite{SMPLX2019} to generate animatable digital characters but sacrifices appearance quality.
There is always a trade-off between static quality and dynamic animation within current methods.
We attribute it to two prominent drawbacks: (1) \textbf{Limitations in representation}: the animatable head prior model struggles to model high-quality texture and geometry (refer to \cref{fig:comparision-w-tada} and \cref{fig:comparision-w-bergman}); (2) \textbf{Challenges in optimization}: aligning the static representation with the dynamic head prior is difficult (refer to \cref{fig:abl-F-SDS}).

In this paper, we propose a novel text-based generation framework, named \textbf{HeadStudio}, by fully exploiting 3D Gaussian splatting (3DGS)~\cite{kerbl3Dgaussians}, which achieves superior rendering quality and real-time performance for novel-view synthesis. 
Our method comprises two components: 
(1) \textbf{Animatable Head Gaussian}: 
We first arm FLAME~\cite{li2017flame}, an animatable head prior model, with 3D Gaussian splatting by rigging each 3D Gaussian point to a mesh.
As an animatable head Gaussian model, we use the head prior model, to deform 3D Gaussians and employ them for high-quality texture and geometry modeling.
(2) \textbf{Text to Avatar Optimization}: 
We enhance the optimization from initialization, distillation, and regularization to jointly learn the shape, texture, and animation, improving the visual appearance and animated quality.
In specific, we introduce super-dense Gaussian initialization to thoroughly cover the head model for faster convergence and improved representation.
To ensure the consistency of the control signal during animation-based training, we denoise the score distillation and utilize the MediaPipe~\cite{lugaresi2019mediapipe} facial landmark map obtained from FLAME as a fine-grained condition for the diffusion model.
To further improve the fidelity of our method, we utilize an adaptive geometry regularization, which gives animatable head Gaussian the ability to employ strict constraints for semantic deformation and represent elements beyond the FLAME space, such as helmets and mustaches simultaneously.

Extensive experiments have shown that HeadStudio is highly effective and superior to state-of-the-art methods in generating dynamic avatars from text~\cite{poole2022dreamfusion, metzer2022latent, zhang2023dreamface, han2023headsculpt, wang2023prolificdreamer, liao2023tada}.
Moreover, our methods can be easily extended to driving generated 3D avatars via both speech-based ~\cite{yi2022generating} and video-based~\cite{DECA:Siggraph2021} methods. 
Overall, our contributions can be summarized as follows.
\begin{itemize}
    \item To the best of our knowledge, we make the first attempt to incorporate 3D Gaussian splatting into the text-based dynamic head avatar generation.
    \item We propose HeadStudio, which arms animatable head prior model with 3DGS and enhances its optimization for creating high-fidelity and animatable head avatars.
    \item HeadStudio is simple, efficient, and effective. With only one end-to-end training stage of 2 hours on 1 NVIDIA A6000 GPU, HeadStudio is able to generate 40 fps high-fidelity head avatars. 
\end{itemize}

\section{Related Work}
\label{sec:related_work}

\textbf{Text-to-2D generation.}
Recently, with the development of vision-language models~\cite{radford2021clip} and diffusion models~\cite{sohl2015deep, ho2020denoising}, great advancements have been made in text-to-image generation (T2I)~\cite{nichol2021glide, ho2022imagen, zhang20232dsurvey}. 
In particular, Stable Diffusion~\cite{stable_diffusion} is a notable framework that trains the diffusion models on latent space, leading to reduced complexity and detail preservation.
With the emergence of text-to-2D models, more applications have been developed~\cite{ma2023vistallama, yang2024doraemongpt, zhang2024prompt}, such as spatial control~\cite{voynov2023sketch, zhang2023controlnet, zhou2024migc}, concept control~\cite{gal2022image, ruiz2022dreambooth, liang2024caphuman}, and image editing~\cite{brooks2023instructpix2pix}.

\noindent\textbf{Text-to-3D generation.}
The success of the 2D generation is incredible.
However, directly transferring the image diffusion models to 3D is challenging, due to the difficulty of 3D data collection.
Recently, Neural Radiance Fields (NeRF)~\cite{mildenhall2020nerf, barron2022mip} opened a new insight for the 3D-aware generation, where only 2D multi-view images are needed in 3D scene reconstruction.
Combining prior knowledge from text-to-2D models, several methods, such as DreamField~\cite{jain2021dreamfields}, DreamFusion~\cite{poole2022dreamfusion}, and SJC~\cite{wang2022sjc}, have been proposed to generate 3D objects guided by text prompt~\cite{li20233dsurvey, zhuo2024vividdreamer}.
Moreover, the recent advancement of text-to-3D generation also inspired multiple applications, including text-guided scenes generation~\cite{cohen2023set, hollein2023text2room}, text-guided 3D editing~\cite{Ayaan2023instructnerf, kamata2023instruct3d}, and text-guided avatar generation~\cite{cao2023dreamavatar, jiang2023avatarcraft, xu2023seeavatar, ma2024x}.

\noindent\textbf{3D Head Generation and Animation.}
Previous 3D head generation is primarily based on statistical models, such as 3DMM~\cite{Blanz19993dmm} and FLAME~\cite{li2017flame}, while current methods utilize 3D-aware Generative Adversarial Networks (GANs)~\cite{schwarz2020graf, chan2021pigan, Chan2022eg3d, An2023panohead, zhang2023multi, shen2024controllable, Portrait3D_sig24}. 
Benefiting from advancements in dynamic scene representation~\cite{Gao-ICCV-DynNeRF, kplanes_2023, Cao2023HEXPLANE}, animatable head avatars reconstruction has been improved.
Given a monocular video or multi-view videos, these methods~\cite{zheng2022imavatar, INSTA:CVPR2023, Zheng2023pointavatar, xu2023avatarmav, qian2023gaussianavatars, kirschstein2023diffusionavatars} reconstruct a photo-realistic head avatar, and animate it based on FLAME.
Specifically, our method was inspired by the technique~\cite{INSTA:CVPR2023, qian2023gaussianavatars} of deforming 3D points through rigging with FLAME mesh.
We enhance its deformation and optimization to adapt to score distillation-based learning.
On the other hand, the text-to-static head avatar methods~\cite{wang2022rodin, zhang2023dreamface, han2023headsculpt, liu2023headartist} show superiority in convenience and generalization.
These methods demonstrate impressive texture and geometry, but are not animatable, limiting their practical application.
Furthermore, TADA~\cite{liao2023tada} and Bergman~\etal~\cite{bergman2023articulated} explore the text-to-dynamic head avatar generation.
Similarly, we utilize FLAME to animate the head avatar, but we use 3DGS to model texture instead of the UV-map.

\begin{figure*}[t]
    \centering
    \includegraphics[width=1.0\textwidth]{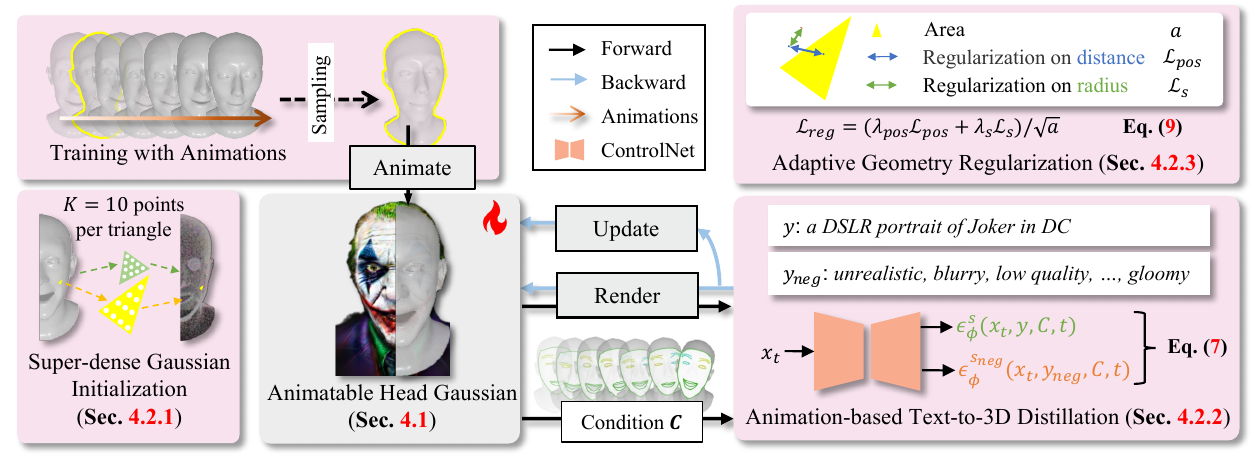}
    \caption{
    Framework of HeadStudio, which integrates animatable head prior model into 3D Gaussian splatting and score distillation sampling. 
    1) \textbf{Animatable Head Gaussian}: each 3D point is rigged to a mesh, and then rotated, scaled, and translated by the mesh deformation.
    2) \textbf{Text to Avatar Optimization}: enhance the optimization from initialization, distillation and regularization, including: super-dense Gaussian initialization, animation-based text-to-3D distillation, and adaptive geometry regularization. 
    }
    \label{fig:overview}
\end{figure*}

\section{Preliminary}
\label{sec:preliminary}
In this section, we provide a brief overview of text to head avatar generation.
The generation process can be seen as distilling knowledge from a diffusion model $\epsilon_\phi$ into a learnable 3D representation $\theta$.
Given camera poses, the corresponding views of the scene can be rendered as images. 
Subsequently, the distillation method guides the image to align with the text description $y$.

\noindent\textbf{Score Distillation Sampling} has been proposed in DreamFusion~\cite{poole2022dreamfusion}.
For a rendered image $x$ from a 3D representation, SDS introduces random noise $\epsilon$ to $x$ at the $t$ timestep, and then uses a pre-trained diffusion model $\epsilon_\phi$ to predict the added noise.
The SDS loss is defined as the difference between predicted and added noise and its gradient is given by
\begin{equation}
    \nabla_\theta \mathcal{L}_{\mathrm{SDS}} = \mathbb{E}_{t, \epsilon}[w(t) (\epsilon^s_{\phi}(x_t;y,t)-\epsilon)\frac{\partial_x}{\partial_\theta}],
\end{equation}
where $x_t = \alpha_t x_0 + \sigma_t \epsilon$ and $w(t)$ is a weighting function, and $s$ is a pre-defined scalar of classifier-free guidance (CFG)~\cite{ho2022classifier}.
The loss estimates and update direction that follows the score function of the diffusion model to move $x$ to the text description region.

\noindent\textbf{3D Gaussian Splatting} ~\cite{kerbl3Dgaussians} is an efficient 3D representation.
It reconstructs a static scene with anisotropic 3D Gaussians, using paired image and camera pose.
Each point is defined by a covariance matrix $\mSigma$ centered at point $\vmu$:
\begin{equation}
    G(\rvx) = e^{-\frac{1}{2}(\rvx-\vmu)^T\mSigma^{-1}(\rvx-\vmu)}.
\end{equation}

Kerbl~\etal~\cite{kerbl3Dgaussians} construct the semi-definite covariance matrix by defining an ellipse using a scaling matrix $\mS$ and a rotation matrix $\mR$, ensuring that the points have meaningful representations:
\begin{equation}
    \mSigma = \mR\mS\mS^T\mR^T.
\end{equation}
The shape and position of a Gaussian point can be represented by a position vector $\vmu\in\mathbb{R}^3$, a scaling vector $\vs \in \mathbb{R}^3$, and a quaternion $\vq \in \mathbb{R}^4$.
Note that we refer $\mR$ to represent the corresponding rotation matrix.
Meanwhile, each 3D Gaussian point has additional parameters: color $\vc$ and opacity $\valpha$, used for splatting-based rendering.
Therefore, a scene can be represented by 3DGS as $\theta_{3DGS} = \left\{ \vmu, \vs, \vq, \vc, \valpha \right\}$.
Given a camera view, the scene can be rendered by the 2D projection of Gaussians via a differentiable tile rasterizer.
In optimization, the gradient of Gaussians is utilized to guide the densification and prune of Gaussians.
We refer readers to \cite{kerbl3Dgaussians, chen2024survey} for more details.

\section{Method}
\label{sec:method}
HeadStudio is a text-to-dynamic head avatar geneartion method.
The created head avatars can be animated by text, speech, and video.
As illustrated in \cref{fig:overview}, the generation pipeline has two key components, including 
(1) the animatable head Gaussian in \cref{sec:F-3DGS}, and (2) text to avatar optimization in \cref{sec:F-SDS}.
Implementation details are discussed in \cref{sec:details}.

\subsection{Animatable Head Gaussian}
\label{sec:F-3DGS}

\noindent\textbf{Animatable Head Prior Model.}
FLAME~\cite{li2017flame} is a vertex-based linear blend skinning (LBS) model, with $N = 5023$ vertices and $4$ joints (neck, jaw, and eyeballs).
The head animation can be formulated by a function:
\begin{equation}
    M(\vbeta, \vgamma, \vpsi): \mathbb{R}^{|\vbeta| \times |\vgamma| \times |\vpsi|} \rightarrow \mathbb{R}^{3N},
\label{eq:flame}
\end{equation}
where $\vbeta\in\mathbb{R}^{|\vbeta|}$, $\vgamma \in\mathbb{R}^{|\vgamma|}$ and $\vpsi\in\mathbb{R}^{|\vpsi|}$ are the shape, pose and expression parameters, respectively (we refer readers to ~\cite{SMPL2015, li2017flame} for the blendshape details).

Recent works have successfully achieved semantic alignment between FLAME and various modalities, such as speech~\cite{yi2022generating, he2023speech4mesh} and talking videos~\cite{DECA:Siggraph2021, MICA:ECCV2022}.
Therefore, existing text-to-dynamic avatar generation methods~\cite{liao2023tada, bergman2023articulated} commonly choose FLAME~\cite{li2017flame} as the base model.
As a result, the created avatars can be semantically animated.
However, the mesh number of FLAME is struggled to model complex textures.
For example, Bergman \etal~\cite{bergman2023articulated} learns one color for each mesh.
It inspires us to arm FLAME with 3D Gaussian points~\cite{kerbl3Dgaussians} for high-quality texture modeling.

\noindent\textbf{Deformable Gaussian Texture.}
To mitigate the limitations of animatable head prior model, we use 3D Gaussian points to model the texture.
The key point is to make sure these points can be deformed semantically by the head prior model.
Following Qian~\etal~\cite{qian2023gaussianavatars}, we assume every 3D Gaussian point is connected with a FLAME mesh.
The FLAME mesh moves and deforms the corresponding points.
Given pose and expression, the FLAME mesh can be calculated by \cref{eq:flame}.
Then, we quantify the mesh triangle by its center position $\vt$, rotation matrix $\Tilde{\mR}$ and area $a$, which describe the triangle's location, orientation and scaling in world space, respectively.
Among them, the rotation matrix is a concatenation of one edge vector, the normal vector of the triangle, and their cross-product.
Formally, we deform the corresponding 3D Gaussian point as
\begin{equation}
    \mR' = \Tilde{\mR}\mR, \qquad \vmu' = \sqrt{a}\Tilde{\mR}\vmu + \vt, \qquad \vs' = \sqrt{a}\vs, 
    \label{eq:f-3dgs}
\end{equation}
where $\vmu'$, $\vs'$ and $\mR'$ are the position vector, scaling vector and rotation matrix of the deformed Gaussian for rendering.
Intuitively, the 3D Gaussian point will be rotated, scaled, and translated by the mesh triangle.
In this way, Gaussians can be seen as a residual term of FLAME to represent intricate geometry and texture.
As a result, FLAME enables the 3DGS to animate semantically, while 3DGS improves the texture representation and rendering efficiency of FLAME.

\noindent\textbf{Joint Learning of Shape, Texture, Animation.}
The intricate texture can be modeled by the deformable Gaussian texture $\theta_{\mathrm{3DGS}}$.
Besides, we assume the shape of head prior model $\theta_{\mathrm{FLAME}} = \left\{\vbeta\right\}$ is learnable.
The learnable shape allows for modeling character more precisely. 
For example, characters like the Hulk in Marvel have larger heads, whereas characters like Elsa in Frozen have thinner cheeks. 
Meanwhile, we notice that excessive shape updates can negatively impact the learning process of 3DGS due to deformation changes. 
Thus, we stop the shape update after a certain number of training steps to ensure stable learning of 3DGS.
As a result, a head avatar can be represented by an animatable head Gaussian as $\theta = \theta_{\mathrm{FLAME}} \cup \theta_{\mathrm{3DGS}}$.

\subsection{Text to Avatar Optimization}
\label{sec:F-SDS}
To jointly learn the shape, texture, and animation of an animatable head Gaussian, we enhance its optimization from initialization, distillation, and regularization, respectively.

\noindent\textbf{Super-dense Gaussian Initialization.}
The supervision signal of SDS loss~\cite{poole2022dreamfusion} in head avatar generation is sparse.
It inspires us to initialize 3D Gaussians that thoroughly cover the head model for faster convergence and improved representation.
In specific, each mesh triangle is initialized with $K$ evenly distributed points.
The positions of the deformed 3D Gaussians $\vmu'$ are calculated by sampling on the FLAME model (with standard pose), with all mesh triangles sharing the same sampling weight.
The deformed scaling $\vs'$ is the square root of the mean distance of its K-nearest neighbor points.
Then, we initialize the position and scaling of 3D Gaussians by the inversion of \cref{eq:f-3dgs}: $\vmu_{init} = \Tilde{\mR}^{-1}((\vmu' - \vt) / \sqrt{a}); \vs_{init} = \vs'/ \sqrt{a}$.
The other learnable parameters in $\theta_{3DGS}$ are initialized following vanilla 3DGS~\cite{kerbl3Dgaussians}.

\noindent\textbf{Animation-based Text-to-3D Distillation.}
The vanilla text-to-3D distillation~\cite{poole2022dreamfusion} produces satisfactory performance in static but falls short in animation.
We attribute it to the absence of new poses and expressions in training.
Therefore, we design a new text-to-3D distillation that adapts to animation.

\noindent\textit{Training with Animations.}
We first incorporate the new pose and expression into training~\cite{liao2023tada, zhang2023avatarstudio}. 
Specifically, we sample pose and expression from real-world motion sequences, such as TalkSHOW~\cite{yi2022generating}, to ensure that the avatar satisfies the textual prompts with a diverse range of animation.

\noindent\textit{FLAME-based Control Generation.}
Training with animations is crucial for dynamic avatar generation.
However, the direct introduction of new pose and expression results in Janus (multi-faces) problem~\cite{hong2023debiasing}, due to the data bias in the diffusion model.
This issue, represented as portrait bias with front-view, straight-looking, and closed mouths,  hinders its application in animation-based distillation.
To address this issue, we introduce the MediaPipe~\cite{lugaresi2019mediapipe} facial landmark map $C$, a fine-grained control signal marking the regions of upper lips, lower lips, eye boundary, eyeballs, and facial boundary~\cite{han2023headsculpt, liu2023headartist}, for more precise and detailed guidance.
It can be extracted from an animatable head Gaussian, which ensures that the control signal aligns well with the Gaussian points when the shape, pose, and expression change.
The loss gradient is formulated as:
\begin{equation}
    \nabla_\theta \mathcal{L}_{\mathrm{SDS}} = \mathbb{E}_{t, \epsilon, \vgamma, \vpsi}[w(t) (\epsilon^{s}_{\phi}(x_t;y,C,t)-\epsilon)\frac{\partial_x}{\partial_\theta}].
\label{eq: f-sds}
\end{equation}

\noindent\textit{Denoised Score Distillation.}
According to our experiments, we find the generated avatars have non-detailed and over-smooth textures.
To solve this issue, we consider the distilled score to be noisy~\cite{hertz2023delta, wang2023prolificdreamer, katzir2023noise}.
Hertz~\etal~\cite{hertz2023delta} indicates that the score can be seen as the noise when the rendered image matches the textual prompt. 
Following NFSD~\cite{katzir2023noise}, we assume the score with a large timestep $t \geq 200$ is noisy, and the rendered image can be seen as matching the negative textural prompts, such as $y_{\mathrm{neg}} = $ ``\textit{unrealistic, blurry, low quality, out of focus, ugly, low contrast, dull, dark, low-resolution, gloomy}''.
Besides, the score with a small timestep $t < 200$ is relatively clean.
As a result, we reorganize the SDS into a piece-wise function:
\begin{equation}
    \nabla_\theta \mathcal{L}_{\mathrm{SDS}} = 
    \begin{cases}
        \mathbb{E}_{t, \epsilon, \vgamma, \vpsi}[w(t) \epsilon^{s}_{\phi}(x_t;y,C,t)\frac{\partial_x}{\partial_\theta}],& {\mathrm{t} < 200}, \\
        \mathbb{E}_{t, \epsilon, \vgamma, \vpsi}[w(t) (\epsilon^{s}_{\phi}(x_t;y,C,t)-\epsilon^{s_{neg}}_{\phi}(x_t;y_{\mathrm{neg}},C,t))\frac{\partial_x}{\partial_\theta}],& \mathrm{t} \geq 200,
    \end{cases}
\end{equation}
where $s_{neg}$ is a pre-defined CFG scalar for negative textual prompts.
Intuitively, we get a cleaner score to improve the avatar's texture.

\noindent\textbf{Adaptive Geometry Regularization.}
To deform semantically, the 3D Gaussians should closely align with the rigged mesh triangle.
Introducing a regularization term for the 3D Gaussians, such as ${\Vert \vmu \Vert}_2$, will lead to the 3D Gaussians being overly concentrated around the mesh center.
Thus, the regularization should inversely scale with the triangle size.
For instance, in the eye and mouth region, where the mesh triangle is small, the rigged Gaussians should have a relatively small scaling and position.
Following Qian~\etal~\cite{qian2023gaussianavatars}, we introduce the position and scaling regularization.
For each triangle, we initially calculate the maximum distance among its center $\vt$ and three vertices, termed as $\tau$, to describe the triangle size.
Then, we formulate the regularization term as:
\begin{equation}
    \mathcal{L}_{\mathrm{pos}} = {\Vert \max({\Vert \sqrt{a}\mR'\vmu \Vert}_2, \tau_{\mathrm{pos}}) \Vert}_2, \qquad
    \mathcal{L}_{\mathrm{s}} = {\Vert \max(\sqrt{a}\vs, \tau_{\mathrm{s}}) \Vert}_{2},
    \label{eq:reg-naive}
\end{equation}
where $\tau_{\mathrm{pos}} = 0.5\tau$ and $\tau_{\mathrm{s}} = 0.5\tau$ are the experimental position tolerance and scaling tolerance, respectively.

\begin{wrapfigure}{r}{2.6cm}
\centering
\includegraphics[width=0.25\textwidth]{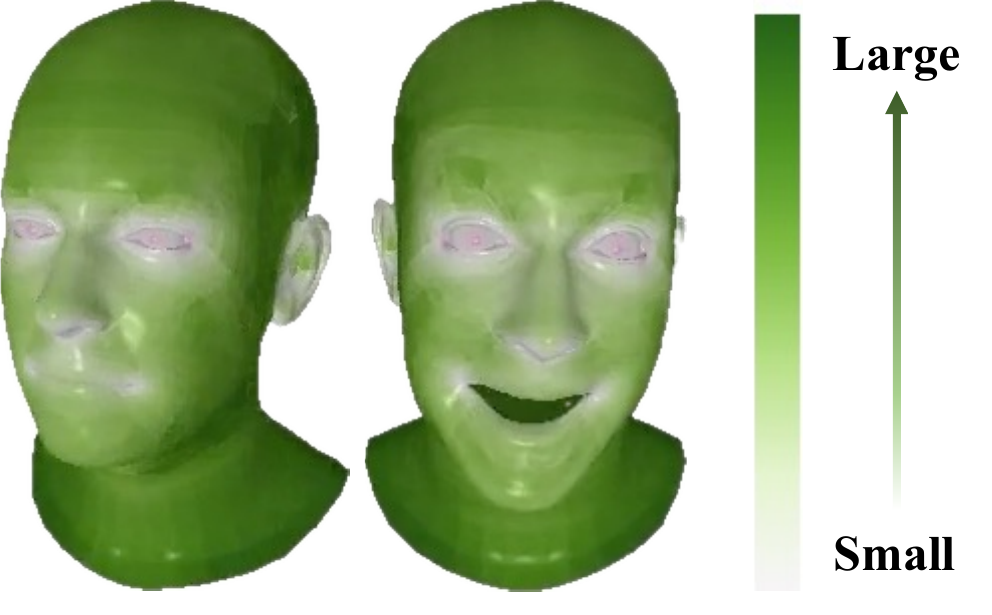}
\caption{
Visualization of Mesh Area.
}
\label{fig:abl-area}
\end{wrapfigure}
The regularization term effectively aligns 3D Gaussians with FLAME. 
It ensures that the 3D Gaussians are positioned around the mesh triangle and can be semantically deformed. 
However, it also restricts animatble head Gaussian from modeling elements outside the space of FLAME in some cases, such as Thor's helmet and Kratos's long mustache, which are essential parts of their identities. 
On the other hand, as shown in \cref{fig:abl-area}, we observe that these elements are located on mesh triangles with large areas.
This observation inspires us to introduce the area $a$ as an adaptive factor:
\begin{equation}
    \mathcal{L}_{\mathrm{reg}} = (\lambda_{\mathrm{pos}}\mathcal{L}_{\mathrm{pos}} +  \lambda_{\mathrm{s}}\mathcal{L}_{\mathrm{s}})/ \sqrt{a}, 
\label{eq:reg}
\end{equation}
where $\lambda_{\mathrm{pos}} = 0.1$ and $\lambda_{\mathrm{s}} = 0.1$.
Through regularization, the avatar demonstrates its ability for semantic deformation and modeling complex appearance.

\begin{figure*}[ht]
    \centering
    \includegraphics[width=1.0\textwidth]{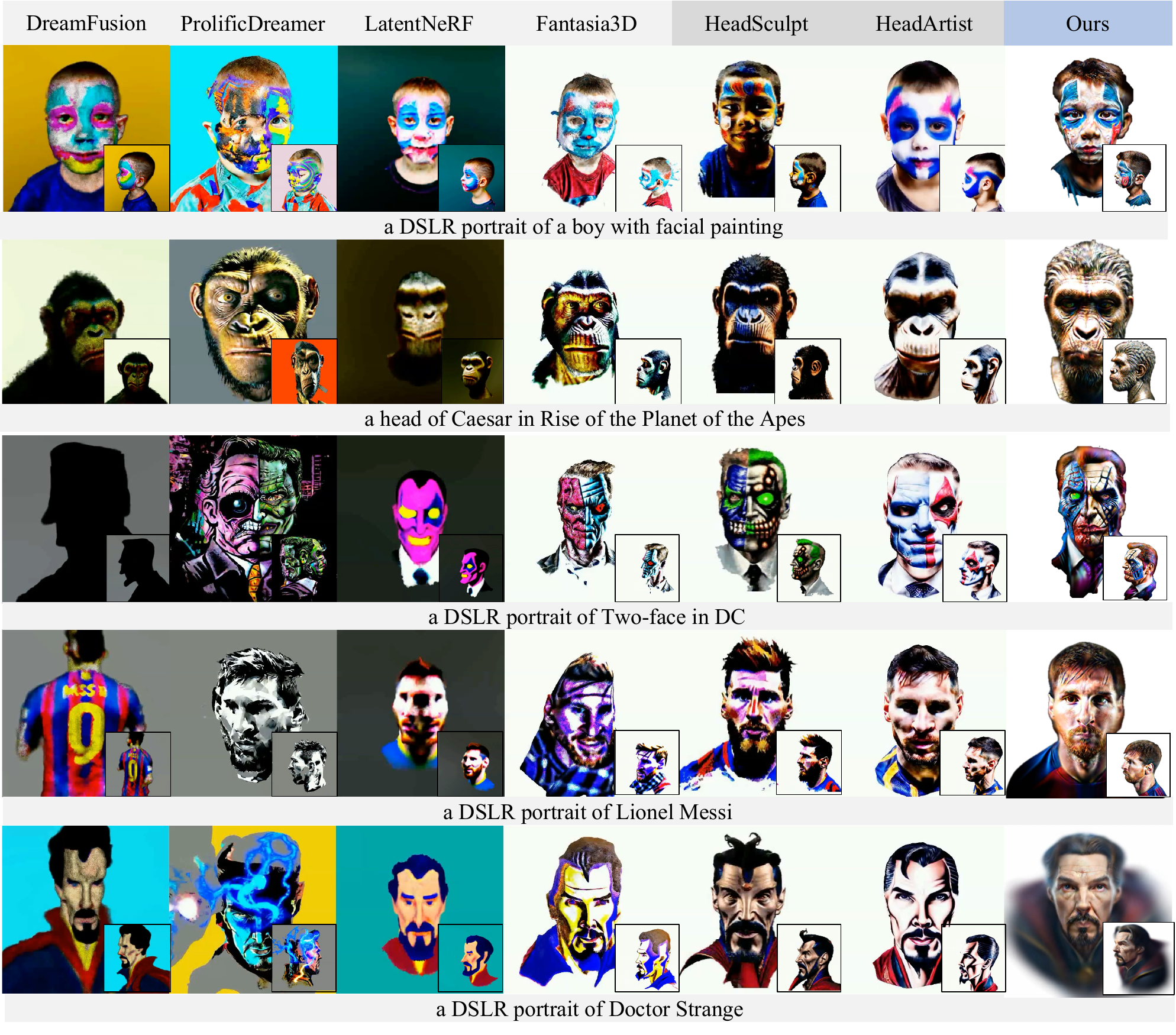}
    \caption{
    Comparison with the text-to-static avatar generation methods.
    Our approach excels at producing high-fidelity head avatars, yielding superior results.
    }
    \label{fig:comparison-w-static-avatar}
\end{figure*}

\begin{figure*}[t]
    \centering
    \includegraphics[width=1.0\textwidth]{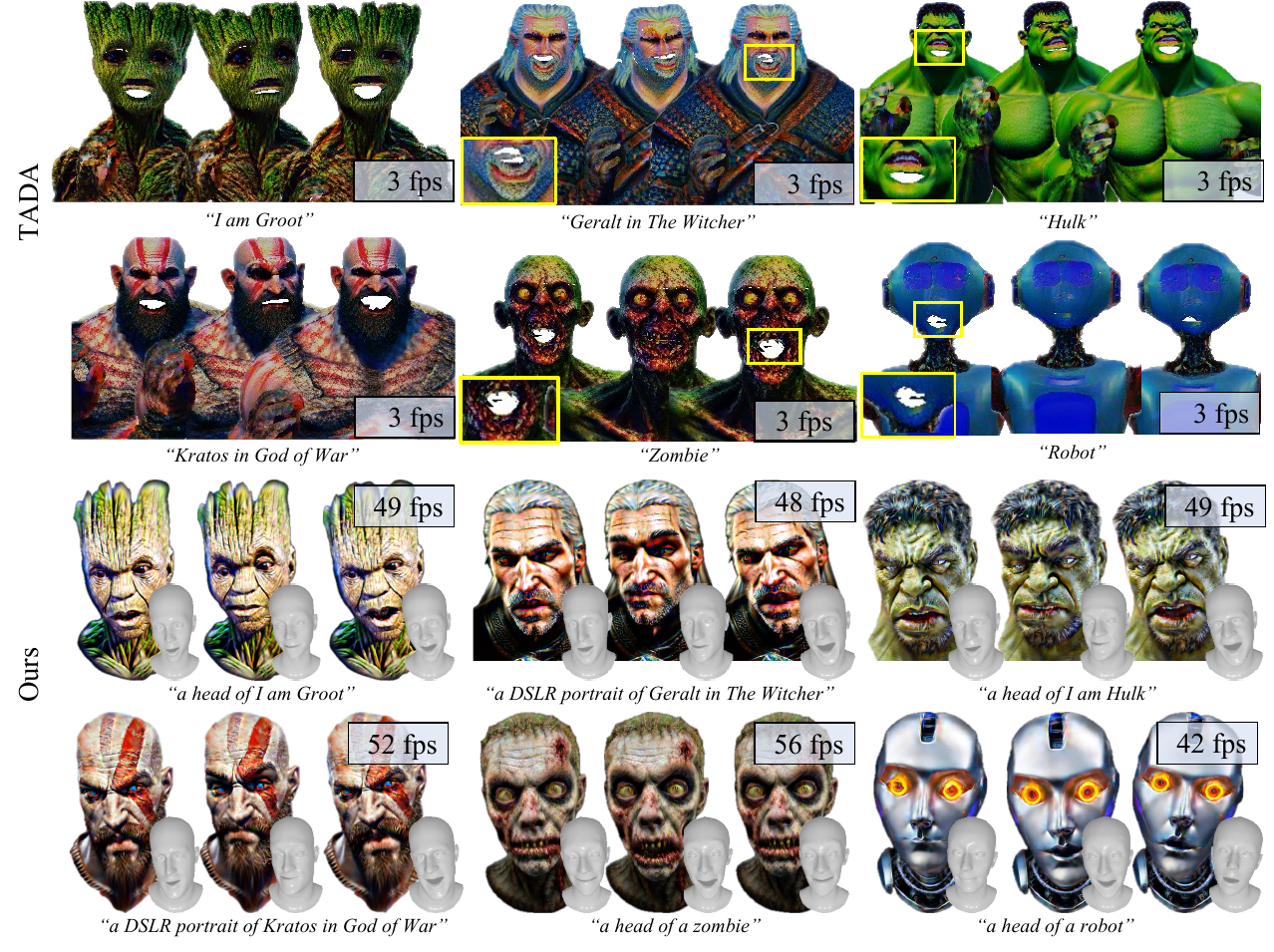}
    \caption{
    Comparison with the text-to-dynamic avatar generation method TADA~\cite{liao2023tada} in terms of semantic alignment and rendering speed.
    The yellow circles indicate semantic misalignment in the mouths, resulting in misplaced mouth texture.
    The rendering speed evaluation on the same device is reported in the blue box.
    The FLAME mesh of the avatar is visualized on the bottom right.
    Our method provides effective semantic alignment, smooth expression deformation, and real-time rendering.
    }
    \label{fig:comparision-w-tada}
\end{figure*}

\begin{figure}[t]
    \centering
    \includegraphics[width=\linewidth]{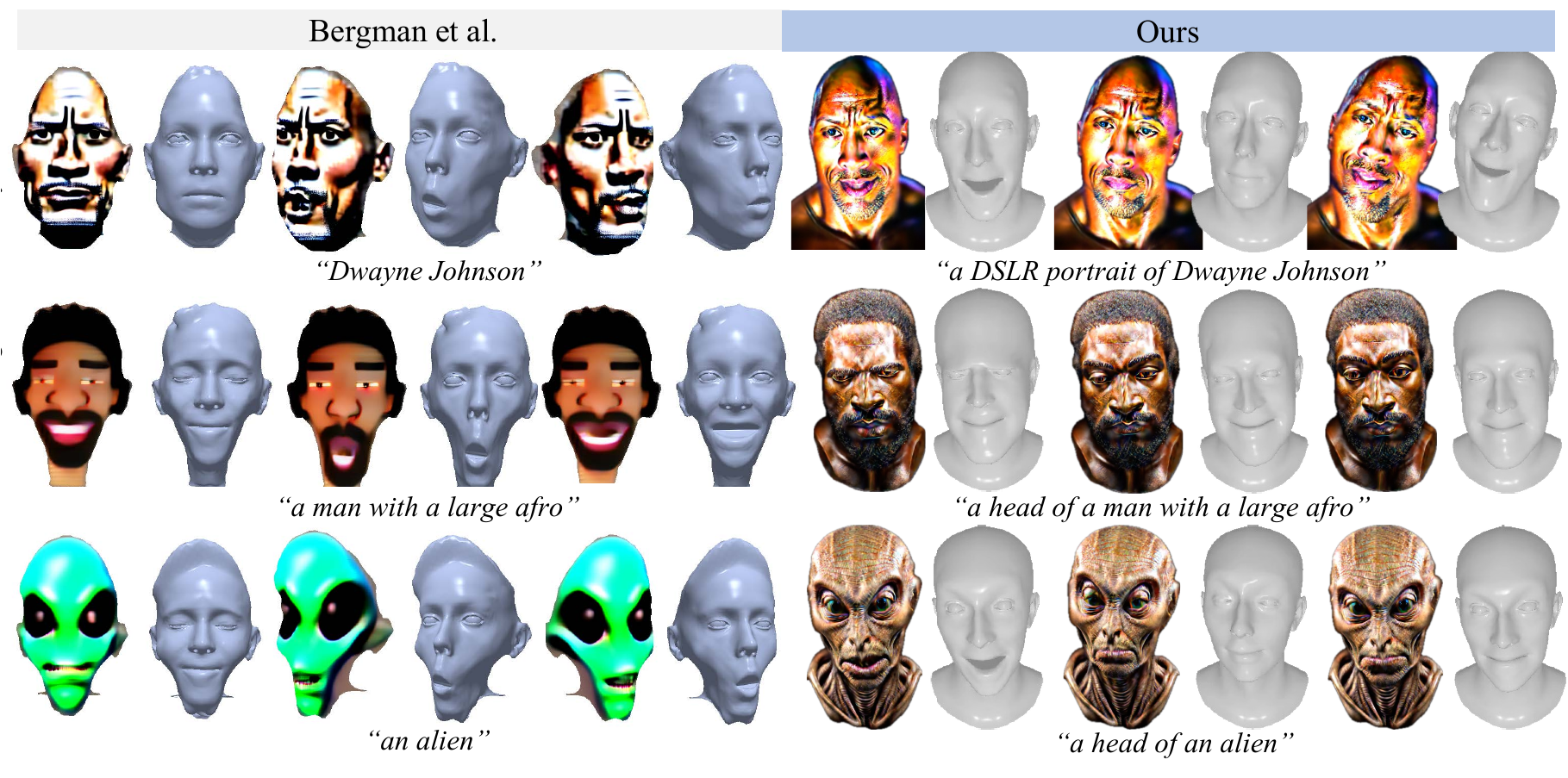}
    \caption{
    Comparison with the text-to-dynamic avatar generation method, Bergman~\etal~\cite{bergman2023articulated}.
    The FLAME mesh of the avatar is visualized on the bottom right.
    Our method demonstrates superior appearance and geometric modeling.
    }
    \label{fig:comparision-w-bergman}
\end{figure}


\subsection{Implementation Details}
\label{sec:details}
\noindent\textbf{Animatable Head Gaussian Details.}
In 3DGS, Kerbl~\etal~\cite{kerbl3Dgaussians} employs a gradient threshold to filter points that require densification.
Nevertheless, the original design cannot handle textual prompts with varying gradient responses.
To address this, we utilize a normalized gradient to identify the points with consistent and significant gradient responses. 
Furthermore, the cloned and split points will inherit the same mesh triangle correspondence of their parent~\cite{qian2023gaussianavatars}.
The densification and pruning iterations setting are following~\cite{liu2023humangaussian}. 
The FLAME's shape size is $|\vgamma|=300$, the expression size is $|\vpsi|=100$ and the pose size is $|\vgamma| = 3 \times 4$ (neck, jaw, left eyeball and right eyeball).

\noindent\textbf{Text to Avatar Optimization Details.}
We initialize animatable head Gaussian with $K = 10$ per triangle.
Besides, we commonly set $s=7.5$ and $s_{neg}=1$ in animation-based text-to-3D distillation~\cite{katzir2023noise}.
In our experiment, we default to using Realistic Vision 5.1 (RV5.1)~\cite{RV5.1} and ControlNetMediaPipeFace~\cite{zhang2023controlnet, ControlNetMediaPipeFace}.
To alleviate the multi-face Janus problem, we also use the view-dependent prompts~\cite{hong2023debiasing}.

\noindent\textbf{Training Details.}
The framework is implemented in PyTorch and threestudio~\cite{threestudio2023}.
We employ a random camera sampling strategy with camera distance range of $\left[1.5, 2.0\right]$, a fovy range of $\left[40^\circ, 70^\circ\right]$, an elevation range of $\left[-30^\circ, 30^\circ\right]$, and an azimuth range of $\left[-180^\circ, 180^\circ\right]$.
We train head avatars with a resolution of 1024 and a batch size of 8. 
The entire training consists of 10,000 iterations.
The overall framework is trained using the Adam optimizer~\cite{kingma2014adam}, with betas of $\left[0.9, 0.99\right]$, 
and learning rates of 5e-5, 1e-3, 1e-2, 1.25e-2, 1e-2, and 1e-3 for mean position $\vmu$, scaling factor $vs$, rotation quaternion $\vq$, color $\vc$, opacity $\valpha$, and FLAME shape $\vbeta$, respectively~\cite{liu2023humangaussian}.
Note that we stop the FLAME shape optimization after 8,000 iterations.
The entire optimization process takes around two hours on a single NVIDIA A6000 (48GB) GPU.

\section{Experiment}
\label{sec:exp}

\begin{table}[!t]
\caption{
\textbf{Quantitative Evaluation.}
Evaluating the coherence of generations with their caption using different CLIP models.
}
\setlength{\tabcolsep}{15pt}
\begin{tabular}{l|ccc}
\hline
CLIP-Score & ViT-L/14$\uparrow$  & ViT-B/16 $\uparrow$ & ViT-B/32 $\uparrow$ \\
\hline
DreamFusion~\cite{poole2022dreamfusion} & 0.244 & 0.302 & 0.300 \\
LatentNeRF~\cite{metzer2022latent} & 0.248 & 0.299 & 0.303 \\
Fantasia3D~\cite{Chen2023fantasia3D} & 0.267 & 0.304 & 0.300 \\
ProlificDreamer~\cite{wang2023prolificdreamer} & 0.268 & 0.320 & 0.308 \\
\hline
HeadSculpt~\cite{han2023headsculpt} & 0.264 & 0.306 & 0.305 \\
HeadArtist~\cite{liu2023headartist} & 0.272 & 0.318 & 0.313 \\
\hline
Ours & \textbf{0.275} & \textbf{0.322} & \textbf{0.317} \\
\hline
\end{tabular}
\label{tab:quan}
\end{table}

\begin{figure}[t]
    \centering
    \includegraphics[width=1.0\linewidth]{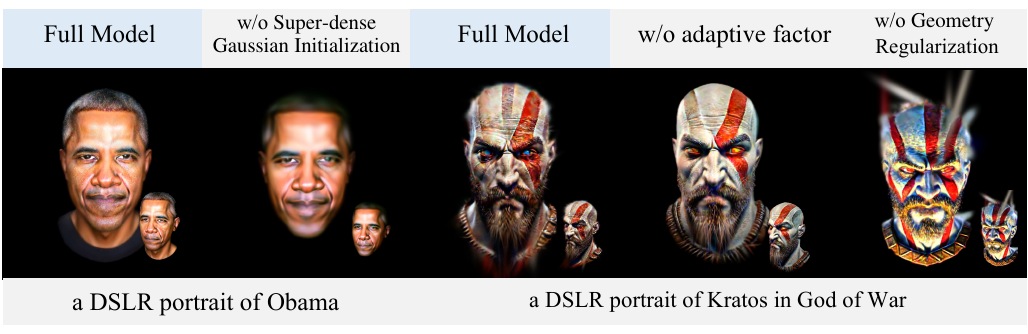}
    \caption{
    \textbf{Ablation Study of Super-dense Gaussian Initialization and Adaptive Geometry Regularization.}
    Super-dense Gaussian initialization enhances the representation ability.
    Geometry regularization imposes a strong restriction to reduce the outline points.
    The adaptive factor in geometry regularization balances restriction and expressiveness.
    }
    \label{fig:abl-F-3DGS}
\end{figure}

\begin{figure}[h]
    \centering
    \includegraphics[width=1.0\linewidth]{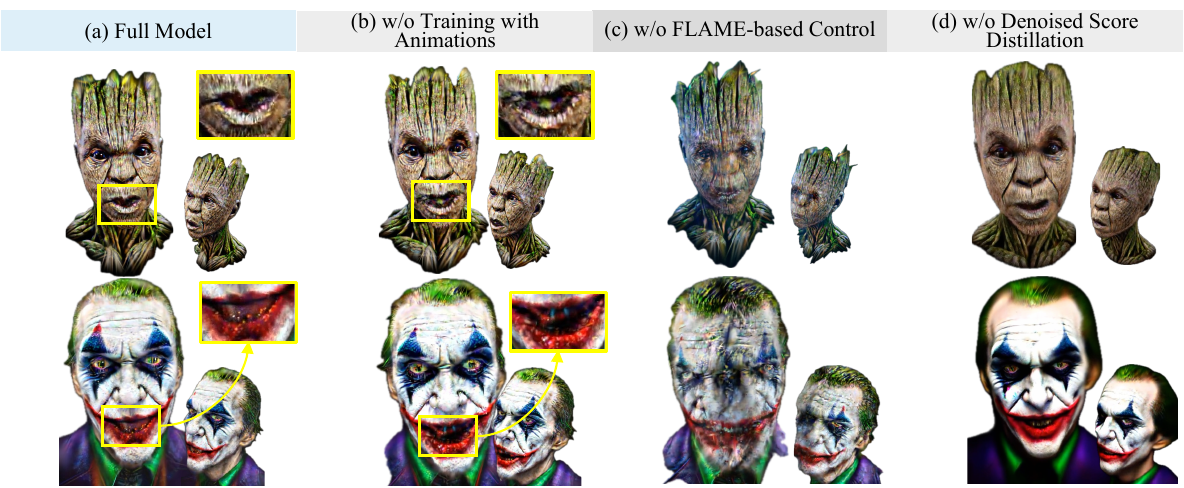}
    \caption{
    \textbf{Ablation Study of Animation-based Text-to-3D Distillation.}
    We investigate the effects of training with animation, FLAME-based control, and denoised score distillation.
    These approaches are dedicated to improving the semantic accuracy of score distillation.
    As a result, animation-based text-to-3D distillation achieves an effective alignment, leading to an accurate expression deformation.
    }
    \label{fig:abl-F-SDS}
\end{figure}

\noindent\textbf{Evaluation.}
We evaluate the quality of head avatars with two settings.
1) \textit{static head avatars}: producing a diverse range of avatars based on various text prompts. 
2) \textit{dynamic head avatars}: driving an avatar with FLAME sequences sampled in TalkSHOW~\cite{yi2022generating}.

\noindent\textbf{Baselines.}
We compare our method with state-of-the-art methods in two settings.
1) \textit{static head avatars}: 
We compare the generation results with six baselines: DreamFusion~\cite{poole2022dreamfusion}, LatentNeRF~\cite{metzer2022latent}, Fantasia3D~\cite{Chen2023fantasia3D} and ProlificDreamer~\cite{wang2023prolificdreamer}, HeadSculpt~\cite{han2023headsculpt} and HeadArtist~\cite{liu2023headartist}.
It is worth noting that HeadSculpt~\cite{han2023headsculpt} and HeadArtist~\cite{liu2023headartist} specialize in text-to-static head avatar generation.
2) \textit{dynamic head avatars}:
We evaluate the efficacy of avatar animation by comparing it with TADA~\cite{liao2023tada} and Bergman~\etal~\cite{bergman2023articulated}. 
Both approaches are based on FLAME and utilize it for animation.

\subsection{Head Avatar Generation}
\noindent\textbf{Static Head Avatar Generation.}
We evaluate the avatar generation quality in terms of geometry and texture.
In \cref{fig:comparison-w-static-avatar}, we evaluate the geometry through novel-view synthesis.
Comparatively, the head-specialized methods produce avatars with superior geometry compared to the text-to-3D methods~\cite{poole2022dreamfusion, metzer2022latent, Chen2023fantasia3D, wang2023prolificdreamer}.
This improvement can be attributed to the integration of FLAME, a reliable head structure prior, which mitigates the multi-face Janus problem~\cite{hong2023debiasing} and enhances the geometry.

On the other hand, we evaluate the texture through quantitative experiments using the CLIP score~\cite{hessel2021clipscore}.
This metric measures the similarity between the given textual prompt and the generated avatars.
A higher CLIP score indicates a closer match between the generated avatar and the text, highlighting a more faithful texture.
Following Liu~\etal~\cite{liu2023headartist}, we report the average CLIP score of 10 text prompts.
\cref{tab:quan} demonstrates that HeadStudio outperforms other methods in three different CLIP variants~\cite{radford2021clip}.
Overall, HeadStudio excels at producing high-fidelity head avatars, outperforming the state-of-the-art text-based methods.

\noindent\textbf{Dynamic Head Avatar Generation.}
We evaluate the efficiency of animation in terms of semantic alignment and rendering speed.
For the evaluation of semantic alignment, we visually represent the talking head sequences, which are controlled by speech~\cite{yi2022generating}.
In \cref{fig:comparision-w-tada}, we compare HeadStudio with TADA~\cite{liao2023tada}.
The yellow circles in the first row indicate a lack of semantic alignment in the mouths of Hulk and Geralt, resulting in misplaced mouth texture.
Our approach achieves excellent semantic alignment and smooth expression deformation.
On the other hand, our method enables real-time rendering.
When compared to TADA, such as Kratos (52 fps \emph{vs}\onedot 3 fps), our method demonstrates its potential in augmented or virtual reality applications.
Furthermore, the comparison in \cref{fig:comparision-w-bergman} indicates the semantic alignment of the method proposed by Bergman \etal~\cite{bergman2023articulated}.
But it lacks in terms of its representation of appearance and geometry. 

\subsection{Ablation Study}
We isolate the various contributions and conducted a series of experiments to assess their impact.
In particular, we examine the design of super-dense Gaussian initialization, animation-based text-to-3D distillation, and adaptive geometry regularization.
At last, we discuss the effect of different diffusion models.

\noindent\textbf{Effect of Super-dense Gaussian Initialization.}
In \cref{fig:abl-F-3DGS}, we present the effect of super-dense Gaussian initialization.
Since the SDS supervision signal is sparse, super-dense Gaussian initialization enhances point coverage on the head model, leading to a favorable initialization and improved avatar fidelity.

\noindent\textbf{Effect of Animation-based Text-to-3D Distillation.}
As illustrated in \cref{fig:abl-F-SDS}, we visualize the effect of each component in text to avatar optimization.
Our method shows the improvements in the following three aspects:
1) Shape (a \emph{vs}\onedot c): FLAME offers precise control signals to address multi-face issues, ensuring ID consistency.
2) Texture (a \emph{vs}\onedot d): Denoised score distillation alleviates the over-smoothing problem in texture by eliminating unnecessary gradients.
3) Animation (a \emph{vs}\onedot b): Training with animations is crucial for artifact elimination (highlighted in yellow box) in deformation.

\noindent\textbf{Effect of Adaptive Geometry Regularization.}
In \cref{fig:abl-F-3DGS}, we also present the effect of adaptive geometry regularization.
Firstly, adaptive geometry regularization could reduces the outline points.
Nevertheless, overly strict regularization weaken the representation ability of animatable head Gaussian, such as the beard of Kratos (fourth column in \cref{fig:abl-F-3DGS}).
To address this, we introduce an adaptive scale factor to balance restriction and expressiveness based on the area of mesh triangle.
Consequently, the restriction of Gaussian points rigged on jaw mesh has been reduced, resulting in a lengthier beard for Kratos (third column in \cref{fig:abl-F-3DGS}).

\begin{figure}[t]
    \centering
    \includegraphics[width=1.0\linewidth]{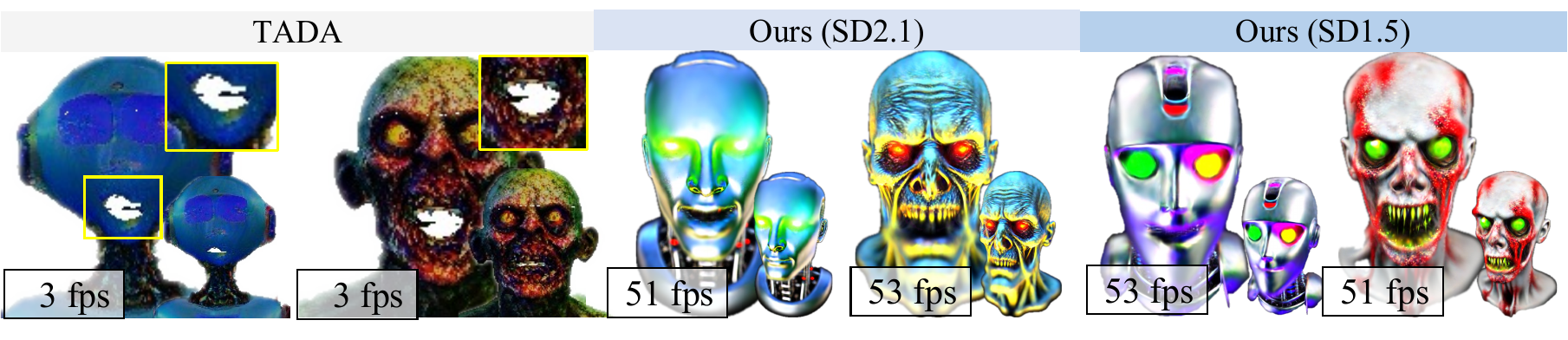}
    \caption{
    \textbf{Ablation Study of Different Diffusion Models.}
    We investigate the effects of different diffusion models, including the Stable Diffusion v2.1 (SD2.1) and Stable Diffusion v1.5 (SD1.5).
    }
    \label{fig:abl-sd}
\end{figure}

\noindent\textbf{Effect of Different Diffusion Models.}
In this paper, we use Realistic Vision 5.1 (RV5.1)~\cite{RV5.1} as the default diffusion model.
Compared to SD2.1~\cite{stable_diffusion} and SD1.5, we observe that RV5.1 is capable of producing head avatars with a more visually appealing appearance.
Meanwhile, we show the results of using SD2.1 (same as TADA~\cite{liao2023tada}) and SD1.5 in \cref{fig:abl-sd}.
HeadStudio can generate avatars with better semantic alignment (texture alignment in mouths) and faster rendering speed (53 fps \emph{vs}\onedot 3 fps) compared with TADA~\cite{liao2023tada}.

\begin{figure}[ht]
    \centering
    \includegraphics[width=1.0\linewidth]{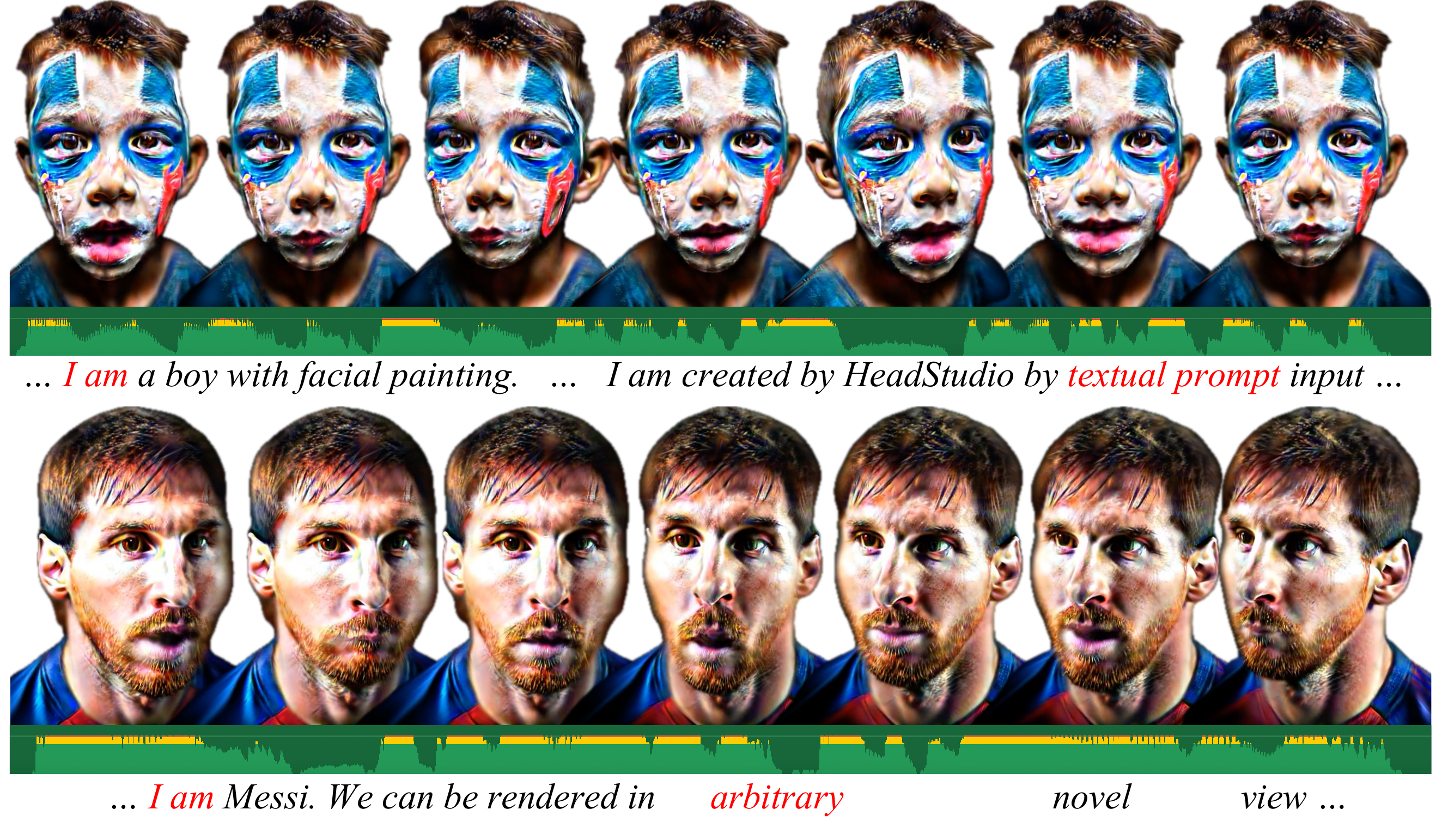}
    \caption{
    \textbf{Application of HeadStudio.}
    We expand our framework by employing TalkSHOW~\cite{yi2022generating} to translate human speech to FLAME sequences.
    From bottom to top: the text input, the corresponding speech clip, and the animated head avatar.
    }
    \label{fig:app-audio}
\end{figure}

\subsection{Application of HeadStudio.}
We further explore the applications of HeadStudio.
\textbf{Audio-based animation} is a widely used technology in conference calls and virtual social presence.
To realize it, we combine our framework with TalkSHOW~\cite{yi2022generating} to translate human speech to FLAME sequences.
\textbf{Text-based animation} can be used for creating talking head videos.
We further expand the audio-based animation framework with a text-to-speech method PlayHT~\cite{PlayHT}.
As shown in \cref{fig:app-audio}, the animation results are semantically aligned with the text input, showing its potential for real-world applications.
We recommend the reader evaluate the performance through the supplementary videos.

\section{Conclusion}
\label{sec:conclusion}
In this paper, we propose HeadStudio, a novel pipeline for generating high-fidelity and animatable 3D head avatars using 3D Gaussian Splatting.
We arm the animatable head prior model with 3DGS for intricate texture and geometry modeling.
Additionally, we enhance its optimization process from initialization, distillation, and regularization to simultaneously learn shape, texture, and animation, resulting in visually pleasing and high-quality animated avatars.
Extensive evaluations demonstrated that our HeadStudio produces high-fidelity and animatble avatars with real-time rendering, outperforming state-of-the-art methods significantly.

\clearpage

\section*{Acknowledgements}
This work was supported in part by the National Key R\&D Program of China under Grant 2022ZD0160101, the National Natural Science Foundation of China (U2336212), the Fundamental Research Funds for the Central Universities (No. 226-2022-00051), the Fundamental Research Funds for the Central Universities (No. 226-2024-00058), the Fundamental Research Funds for the Zhejiang Provincial Universities (No. 226-2024-00208), the “Leading Goose” R\&D Program of Zhejiang Province under Grant 2024C01101, and the China Postdoctoral Science Foundation (524000-X92302).

%
%
\bibliographystyle{splncs04}
\bibliography{main}

\begin{thebibliography}{10}
\providecommand{\url}[1]{\texttt{#1}}
\providecommand{\urlprefix}{URL }
\providecommand{\doi}[1]{https://doi.org/#1}

\bibitem{ControlNetMediaPipeFace}
Controlnetmediapipeface, \url{https://huggingface.co/CrucibleAI/ControlNetMediaPipeFace}

\bibitem{PlayHT}
Playht, \url{https://play.ht/}

\bibitem{RV5.1}
Realistic vision 5.1, \url{https://huggingface.co/stablediffusionapi/realistic-vision-51}

\bibitem{An2023panohead}
An, S., Xu, H., Shi, Y., Song, G., Ogras, U.Y., Luo, L.: Panohead: Geometry-aware 3d full-head synthesis in 360$^{\circ} $. In: Proceedings of the IEEE/CVF Conference on Computer Vision and Pattern Recognition (CVPR). pp. 20950--20959 (June 2023)

\bibitem{barron2022mip}
Barron, J.T., Mildenhall, B., Verbin, D., Srinivasan, P.P., Hedman, P.: Mip-nerf 360: Unbounded anti-aliased neural radiance fields. In: Proceedings of the IEEE/CVF Conference on Computer Vision and Pattern Recognition (CVPR). pp. 5470--5479 (2022)

\bibitem{bergman2023articulated}
Bergman, A.W., Yifan, W., Wetzstein, G.: Articulated 3d head avatar generation using text-to-image diffusion models. arXiv preprint arXiv:2307.04859  (2023)

\bibitem{Blanz19993dmm}
Blanz, V., Vetter, T.: A morphable model for the synthesis of {3D} faces. In: SIGGRAPH (1999). \doi{10.1145/311535.311556}

\bibitem{brooks2023instructpix2pix}
Brooks, T., Holynski, A., Efros, A.A.: Instructpix2pix: Learning to follow image editing instructions. In: Proceedings of the IEEE/CVF Conference on Computer Vision and Pattern Recognition (CVPR). pp. 18392--18402 (2023)

\bibitem{Cao2023HEXPLANE}
Cao, A., Johnson, J.: Hexplane: A fast representation for dynamic scenes. CVPR  (2023)

\bibitem{cao2023dreamavatar}
Cao, Y., Cao, Y.P., Han, K., Shan, Y., Wong, K.Y.K.: Dreamavatar: Text-and-shape guided 3d human avatar generation via diffusion models. arXiv preprint arXiv:2304.00916  (2023)

\bibitem{Chan2022eg3d}
Chan, E.R., Lin, C.Z., Chan, M.A., Nagano, K., Pan, B., Mello, S.D., Gallo, O., Guibas, L., Tremblay, J., Khamis, S., Karras, T., Wetzstein, G.: Efficient geometry-aware {3D} generative adversarial networks. In: Proceedings of the IEEE/CVF Conference on Computer Vision and Pattern Recognition (CVPR) (2022)

\bibitem{chan2021pigan}
Chan, E.R., Monteiro, M., Kellnhofer, P., Wu, J., Wetzstein, G.: pi-gan: Periodic implicit generative adversarial networks for 3d-aware image synthesis. In: Proceedings of the IEEE/CVF conference on computer vision and pattern recognition. pp. 5799--5809 (2021)

\bibitem{chen2024survey}
Chen, G., Wang, W.: A survey on 3d gaussian splatting. arXiv preprint arXiv:2401.03890  (2024)

\bibitem{Chen2023fantasia3D}
Chen, R., Chen, Y., Jiao, N., Jia, K.: Fantasia3d: Disentangling geometry and appearance for high-quality text-to-3d content creation. In: Proceedings of the IEEE/CVF International Conference on Computer Vision (ICCV) (October 2023)

\bibitem{cohen2023set}
Cohen-Bar, D., Richardson, E., Metzer, G., Giryes, R., Cohen-Or, D.: Set-the-scene: Global-local training for generating controllable nerf scenes. arXiv preprint arXiv:2303.13450  (2023)

\bibitem{DECA:Siggraph2021}
Feng, Y., Feng, H., Black, M.J., Bolkart, T.: Learning an animatable detailed {3D} face model from in-the-wild images. ACM Transactions on Graphics, (Proc. SIGGRAPH)  \textbf{40}(8) (2021), \url{https://doi.org/10.1145/3450626.3459936}

\bibitem{kplanes_2023}
Fridovich-Keil, S., Meanti, G., Warburg, F.R., Recht, B., Kanazawa, A.: K-planes: Explicit radiance fields in space, time, and appearance. In: CVPR (2023)

\bibitem{gal2022image}
Gal, R., Alaluf, Y., Atzmon, Y., Patashnik, O., Bermano, A.H., Chechik, G., Cohen-Or, D.: An image is worth one word: Personalizing text-to-image generation using textual inversion. arXiv preprint arXiv:2208.01618  (2022)

\bibitem{Gao-ICCV-DynNeRF}
Gao, C., Saraf, A., Kopf, J., Huang, J.B.: Dynamic view synthesis from dynamic monocular video. In: Proceedings of the IEEE International Conference on Computer Vision (2021)

\bibitem{threestudio2023}
Guo, Y.C., Liu, Y.T., Shao, R., Laforte, C., Voleti, V., Luo, G., Chen, C.H., Zou, Z.X., Wang, C., Cao, Y.P., Zhang, S.H.: threestudio: A unified framework for 3d content generation. \url{https://github.com/threestudio-project/threestudio} (2023)

\bibitem{han2023headsculpt}
Han, X., Cao, Y., Han, K., Zhu, X., Deng, J., Song, Y.Z., Xiang, T., Wong, K.Y.K.: Headsculpt: Crafting 3d head avatars with text. arXiv preprint arXiv:2306.03038  (2023)

\bibitem{Ayaan2023instructnerf}
Haque, A., Tancik, M., Efros, A., Holynski, A., Kanazawa, A.: Instruct-nerf2nerf: Editing 3d scenes with instructions. In: Proceedings of the IEEE/CVF International Conference on Computer Vision (ICCV) (2023)

\bibitem{he2023speech4mesh}
He, S., He, H., Yang, S., Wu, X., Xia, P., Yin, B., Liu, C., Dai, L., Xu, C.: Speech4mesh: Speech-assisted monocular 3d facial reconstruction for speech-driven 3d facial animation. In: Proceedings of the IEEE/CVF International Conference on Computer Vision. pp. 14192--14202 (2023)

\bibitem{hertz2023delta}
Hertz, A., Aberman, K., Cohen-Or, D.: Delta denoising score. In: Proceedings of the IEEE/CVF International Conference on Computer Vision. pp. 2328--2337 (2023)

\bibitem{hessel2021clipscore}
Hessel, J., Holtzman, A., Forbes, M., Bras, R.L., Choi, Y.: Clipscore: A reference-free evaluation metric for image captioning. arXiv preprint arXiv:2104.08718  (2021)

\bibitem{ho2022imagen}
Ho, J., Chan, W., Saharia, C., Whang, J., Gao, R., Gritsenko, A., Kingma, D.P., Poole, B., Norouzi, M., Fleet, D.J., et~al.: Imagen video: High definition video generation with diffusion models. arXiv preprint arXiv:2210.02303  (2022)

\bibitem{ho2020denoising}
Ho, J., Jain, A., Abbeel, P.: Denoising diffusion probabilistic models. Advances in Neural Information Processing Systems (NeurIPS)  \textbf{33},  6840--6851 (2020)

\bibitem{ho2022classifier}
Ho, J., Salimans, T.: Classifier-free diffusion guidance. arXiv preprint arXiv:2207.12598  (2022)

\bibitem{hollein2023text2room}
H{\"o}llein, L., Cao, A., Owens, A., Johnson, J., Nie{\ss}ner, M.: Text2room: Extracting textured 3d meshes from 2d text-to-image models. arXiv preprint arXiv:2303.11989  (2023)

\bibitem{hong2023debiasing}
Hong, S., Ahn, D., Kim, S.: Debiasing scores and prompts of 2d diffusion for robust text-to-3d generation. arXiv preprint arXiv:2303.15413  (2023)

\bibitem{jain2021dreamfields}
Jain, A., Mildenhall, B., Barron, J.T., Abbeel, P., Poole, B.: Zero-shot text-guided object generation with dream fields. In: Proceedings of the IEEE/CVF Conference on Computer Vision and Pattern Recognition (CVPR) (2022)

\bibitem{jiang2023avatarcraft}
Jiang, R., Wang, C., Zhang, J., Chai, M., He, M., Chen, D., Liao, J.: Avatarcraft: Transforming text into neural human avatars with parameterized shape and pose control. arXiv preprint arXiv:2303.17606  (2023)

\bibitem{kamata2023instruct3d}
Kamata, H., Sakuma, Y., Hayakawa, A., Ishii, M., Narihira, T.: Instruct 3d-to-3d: Text instruction guided 3d-to-3d conversion. arXiv preprint arXiv:2303.15780  (2023)

\bibitem{katzir2023noise}
Katzir, O., Patashnik, O., Cohen-Or, D., Lischinski, D.: Noise-free score distillation. arXiv preprint arXiv:2310.17590  (2023)

\bibitem{kerbl3Dgaussians}
Kerbl, B., Kopanas, G., Leimk{\"u}hler, T., Drettakis, G.: 3d gaussian splatting for real-time radiance field rendering. ACM Transactions on Graphics  \textbf{42}(4) (July 2023), \url{https://repo-sam.inria.fr/fungraph/3d-gaussian-splatting/}

\bibitem{kingma2014adam}
Kingma, D.P., Ba, J.: Adam: A method for stochastic optimization. arXiv preprint arXiv:1412.6980  (2014)

\bibitem{kirschstein2023diffusionavatars}
Kirschstein, T., Giebenhain, S., Nie{\ss}ner, M.: Diffusionavatars: Deferred diffusion for high-fidelity 3d head avatars. arXiv preprint arXiv:2311.18635  (2023)

\bibitem{li20233dsurvey}
Li, C., Zhang, C., Waghwase, A., Lee, L.H., Rameau, F., Yang, Y., Bae, S.H., Hong, C.S.: Generative ai meets 3d: A survey on text-to-3d in aigc era. arXiv preprint arXiv:2305.06131  (2023)

\bibitem{li2017flame}
Li, T., Bolkart, T., Black, M.J., Li, H., Romero, J.: Learning a model of facial shape and expression from 4d scans. ACM Trans. Graph.  \textbf{36}(6),  194--1 (2017)

\bibitem{liang2024caphuman}
Liang, C., Ma, F., Zhu, L., Deng, Y., Yang, Y.: Caphuman: Capture your moments in parallel universes. In: Proceedings of the IEEE/CVF Conference on Computer Vision and Pattern Recognition. pp. 6400--6409 (2024)

\bibitem{liao2023tada}
Liao, T., Yi, H., Xiu, Y., Tang, J., Huang, Y., Thies, J., Black, M.J.: Tada! text to animatable digital avatars. arXiv preprint arXiv:2308.10899  (2023)

\bibitem{liu2023headartist}
Liu, H., Wang, X., Wan, Z., Shen, Y., Song, Y., Liao, J., Chen, Q.: Headartist: Text-conditioned 3d head generation with self score distillation. arXiv preprint arXiv:2312.07539  (2023)

\bibitem{liu2023humangaussian}
Liu, X., Zhan, X., Tang, J., Shan, Y., Zeng, G., Lin, D., Liu, X., Liu, Z.: Humangaussian: Text-driven 3d human generation with gaussian splatting. arXiv preprint arXiv:2311.17061  (2023)

\bibitem{SMPL2015}
Loper, M., Mahmood, N., Romero, J., Pons-Moll, G., Black, M.J.: Smpl: A skinned multi-person linear model. ACM Trans. Graph.  \textbf{34}(6),  248:1--248:16 (Oct 2015)

\bibitem{lugaresi2019mediapipe}
Lugaresi, C., Tang, J., Nash, H., McClanahan, C., Uboweja, E., Hays, M., Zhang, F., Chang, C.L., Yong, M.G., Lee, J., et~al.: Mediapipe: A framework for building perception pipelines. arXiv preprint arXiv:1906.08172  (2019)

\bibitem{luo2024gaussianhair}
Luo, H., Ouyang, M., Zhao, Z., Jiang, S., Zhang, L., Zhang, Q., Yang, W., Xu, L., Yu, J.: Gaussianhair: Hair modeling and rendering with light-aware gaussians. arXiv preprint arXiv:2402.10483  (2024)

\bibitem{ma2023vistallama}
Ma, F., Jin, X., Wang, H., Xian, Y., Feng, J., Yang, Y.: Vista-llama: Reliable video narrator via equal distance to visual tokens (2023)

\bibitem{ma2024x}
Ma, Y., Lin, Z., Ji, J., Fan, Y., Sun, X., Ji, R.: X-oscar: A progressive framework for high-quality text-guided 3d animatable avatar generation. arXiv preprint arXiv:2405.00954  (2024)

\bibitem{metzer2022latent}
Metzer, G., Richardson, E., Patashnik, O., Giryes, R., Cohen-Or, D.: Latent-nerf for shape-guided generation of 3d shapes and textures. arXiv preprint arXiv:2211.07600  (2022)

\bibitem{mildenhall2020nerf}
Mildenhall, B., Srinivasan, P.P., Tancik, M., Barron, J.T., Ramamoorthi, R., Ng, R.: Nerf: Representing scenes as neural radiance fields for view synthesis. In: Proceedings of the European Conference on Computer Vision (ECCV) (2020)

\bibitem{nichol2021glide}
Nichol, A., Dhariwal, P., Ramesh, A., Shyam, P., Mishkin, P., McGrew, B., Sutskever, I., Chen, M.: Glide: Towards photorealistic image generation and editing with text-guided diffusion models. arXiv preprint arXiv:2112.10741  (2021)

\bibitem{SMPLX2019}
Pavlakos, G., Choutas, V., Ghorbani, N., Bolkart, T., Osman, A.A.A., Tzionas, D., Black, M.J.: Expressive body capture: 3d hands, face, and body from a single image. In: Proceedings IEEE Conf. on Computer Vision and Pattern Recognition (CVPR). pp. 10975--10985 (Jun 2019), \url{http://smpl-x.is.tue.mpg.de}

\bibitem{poole2022dreamfusion}
Poole, B., Jain, A., Barron, J.T., Mildenhall, B.: Dreamfusion: Text-to-3d using 2d diffusion. arXiv preprint arXiv:2209.14988  (2022)

\bibitem{qian2023gaussianavatars}
Qian, S., Kirschstein, T., Schoneveld, L., Davoli, D., Giebenhain, S., Nie{\ss}ner, M.: Gaussianavatars: Photorealistic head avatars with rigged 3d gaussians. arXiv preprint arXiv:2312.02069  (2023)

\bibitem{radford2021clip}
Radford, A., Kim, J.W., Hallacy, C., Ramesh, A., Goh, G., Agarwal, S., Sastry, G., Askell, A., Mishkin, P., Clark, J., et~al.: Learning transferable visual models from natural language supervision. In: Proceedings of the International Conference on Machine Learning (ICML). pp. 8748--8763 (2021)

\bibitem{stable_diffusion}
Rombach, R., Blattmann, A., Lorenz, D., Esser, P., Ommer, B.: High-resolution image synthesis with latent diffusion models. In: Proceedings of the IEEE/CVF Conference on Computer Vision and Pattern Recognition (CVPR). pp. 10684--10695 (2022)

\bibitem{ruiz2022dreambooth}
Ruiz, N., Li, Y., Jampani, V., Pritch, Y., Rubinstein, M., Aberman, K.: Dreambooth: Fine tuning text-to-image diffusion models for subject-driven generation. arXiv preprint arxiv:2208.12242  (2022)

\bibitem{schwarz2020graf}
Schwarz, K., Liao, Y., Niemeyer, M., Geiger, A.: Graf: Generative radiance fields for 3d-aware image synthesis. Advances in Neural Information Processing Systems  \textbf{33},  20154--20166 (2020)

\bibitem{shen2021deep}
Shen, T., Gao, J., Yin, K., Liu, M.Y., Fidler, S.: Deep marching tetrahedra: a hybrid representation for high-resolution 3d shape synthesis. Advances in Neural Information Processing Systems  \textbf{34},  6087--6101 (2021)

\bibitem{shen2024controllable}
Shen, X., Ma, J., Zhou, C., Yang, Z.: Controllable 3d face generation with conditional style code diffusion. In: Proceedings of the AAAI Conference on Artificial Intelligence. vol.~38, pp. 4811--4819 (2024)

\bibitem{sohl2015deep}
Sohl-Dickstein, J., Weiss, E., Maheswaranathan, N., Ganguli, S.: Deep unsupervised learning using nonequilibrium thermodynamics. In: International Conference on Machine Learning. pp. 2256--2265. PMLR (2015)

\bibitem{voynov2023sketch}
Voynov, A., Aberman, K., Cohen-Or, D.: Sketch-guided text-to-image diffusion models. In: ACM SIGGRAPH 2023 Conference Proceedings. pp. 1--11 (2023)

\bibitem{wang2022sjc}
Wang, H., Du, X., Li, J., Yeh, R.A., Shakhnarovich, G.: Score jacobian chaining: Lifting pretrained 2d diffusion models for 3d generation. In: Proceedings of the IEEE/CVF Conference on Computer Vision and Pattern Recognition (CVPR) (2022)

\bibitem{wang2022rodin}
Wang, T., Zhang, B., Zhang, T., Gu, S., Bao, J., Baltrusaitis, T., Shen, J., Chen, D., Wen, F., Chen, Q., et~al.: Rodin: A generative model for sculpting 3d digital avatars using diffusion. arXiv preprint arXiv:2212.06135  (2022)

\bibitem{wang2023prolificdreamer}
Wang, Z., Lu, C., Wang, Y., Bao, F., Li, C., Su, H., Zhu, J.: Prolificdreamer: High-fidelity and diverse text-to-3d generation with variational score distillation. arXiv preprint arXiv:2305.16213  (2023)

\bibitem{wei2024aniportrait}
Wei, H., Yang, Z., Wang, Z.: Aniportrait: Audio-driven synthesis of photorealistic portrait animations (2024)

\bibitem{Portrait3D_sig24}
Wu, Y., Xu, H., Tang, X., Chen, X., Tang, S., Zhang, Z., Li, C., Jin, X.: Portrait3d: Text-guided high-quality 3d portrait generation using pyramid representation and gans prior. ACM Trans. Graph.  \textbf{43}(4) (Jul 2024). \doi{10.1145/3658162}, \url{https://doi.org/10.1145/3658162}

\bibitem{xu2023seeavatar}
Xu, Y., Yang, Z., Yang, Y.: Seeavatar: Photorealistic text-to-3d avatar generation with constrained geometry and appearance. arXiv preprint arXiv:2312.08889  (2023)

\bibitem{xu2023avatarmav}
Xu, Y., Wang, L., Zhao, X., Zhang, H., Liu, Y.: Avatarmav: Fast 3d head avatar reconstruction using motion-aware neural voxels. In: ACM SIGGRAPH 2023 Conference Proceedings (2023)

\bibitem{yang2024doraemongpt}
Yang, Z., Chen, G., Li, X., Wang, W., Yang, Y.: Doraemongpt: Toward understanding dynamic scenes with large language models (exemplified as a video agent). In: ICML (2024)

\bibitem{yi2022generating}
Yi, H., Liang, H., Liu, Y., Cao, Q., Wen, Y., Bolkart, T., Tao, D., Black, M.J.: Generating holistic 3d human motion from speech. In: CVPR (2023)

\bibitem{zhang20232dsurvey}
Zhang, C., Zhang, C., Zhang, M., Kweon, I.S.: Text-to-image diffusion model in generative ai: A survey. arXiv preprint arXiv:2303.07909  (2023)

\bibitem{zhang2023avatarstudio}
Zhang, J., Zhang, X., Zhang, H., Liew, J.H., Zhang, C., Yang, Y., Feng, J.: Avatarstudio: High-fidelity and animatable 3d avatar creation from text. arXiv preprint arXiv:2311.17917  (2023)

\bibitem{zhang2023dreamface}
Zhang, L., Qiu, Q., Lin, H., Zhang, Q., Shi, C., Yang, W., Shi, Y., Yang, S., Xu, L., Yu, J.: Dreamface: Progressive generation of animatable 3d faces under text guidance. arXiv preprint arXiv:2304.03117  (2023)

\bibitem{zhang2023controlnet}
Zhang, L., Rao, A., Agrawala, M.: Adding conditional control to text-to-image diffusion models (2023)

\bibitem{zhang2023multi}
Zhang, X., Zheng, Z., Gao, D., Zhang, B., Yang, Y., Chua, T.S.: Multi-view consistent generative adversarial networks for compositional 3d-aware image synthesis. International Journal of Computer Vision  \textbf{131}(8),  2219--2242 (2023)

\bibitem{zhang2024prompt}
Zhang, Y., Fan, H., Yang, Y.: Prompt-aware adapter: Towards learning adaptive visual tokens for multimodal large language models. arXiv preprint arXiv:2405.15684  (2024)

\bibitem{zheng2022imavatar}
Zheng, Y., Abrevaya, V.F., Bühler, M.C., Chen, X., Black, M.J., Hilliges, O.: {I} {M} {Avatar}: Implicit morphable head avatars from videos. In: Proceedings of the IEEE/CVF Conference on Computer Vision and Pattern Recognition (CVPR) (2022)

\bibitem{Zheng2023pointavatar}
Zheng, Y., Yifan, W., Wetzstein, G., Black, M.J., Hilliges, O.: Pointavatar: Deformable point-based head avatars from videos. In: Proceedings of the IEEE/CVF Conference on Computer Vision and Pattern Recognition (CVPR) (2023)

\bibitem{zhou2024migc}
Zhou, D., Li, Y., Ma, F., Zhang, X., Yang, Y.: Migc: Multi-instance generation controller for text-to-image synthesis. In: Proceedings of the IEEE/CVF Conference on Computer Vision and Pattern Recognition. pp. 6818--6828 (2024)

\bibitem{zhuo2024vividdreamer}
Zhuo, W., Ma, F., Fan, H., Yang, Y.: Vividdreamer: Invariant score distillation for hyper-realistic text-to-3d generation. In: ECCV (2024)

\bibitem{MICA:ECCV2022}
Zielonka, W., Bolkart, T., Thies, J.: Towards metrical reconstruction of human faces. In: European Conference on Computer Vision (2022)

\bibitem{INSTA:CVPR2023}
Zielonka, W., Bolkart, T., Thies, J.: Instant volumetric head avatars. In: Proceedings of the IEEE/CVF Conference on Computer Vision and Pattern Recognition (CVPR) (2023)

\bibitem{zuffi2017smal}
Zuffi, S., Kanazawa, A., Jacobs, D., Black, M.J.: {3D} menagerie: Modeling the {3D} shape and pose of animals. In: Proceedings of the IEEE/CVF Conference on Computer Vision and Pattern Recognition (CVPR) (Jul 2017)

\end{thebibliography}

\clearpage
\renewcommand\thesection{\Alph{section}}
\setcounter{page}{1}
\setcounter{section}{0}
\setcounter{figure}{10}
\setcounter{table}{1}
\setcounter{equation}{9}

{
\centering
\Large
\textbf{HeadStudio: Text to Animatable Head Avatars with 3D Gaussian Splatting}\\
\vspace{0.5em}Supplementary Material \\
\vspace{1.0em}
}

\section{Additional Implementation Details}

\subsection{Text to Animatable Avatar Optimization}
For each text prompt, we first initialize an animatable head Gaussian via the super-dense Gaussian initialization.
Each iteration of HeadStudio performs the following: 
(1) randomly sample a camera and animation inputs (pose and expression);
(2) drive the animatable head Gaussian with the given pose and expression and render an image from that camera;
(3) compute the gradients of the animation-based text-to-3D distillation;
(4) compute the loss of the adaptive geometry regularization;
At the end of an iteration, we update the animatable head Gaussian parameters using an optimizer.

\begin{figure}[h]
    \centering
    \includegraphics[width=1.0\linewidth]{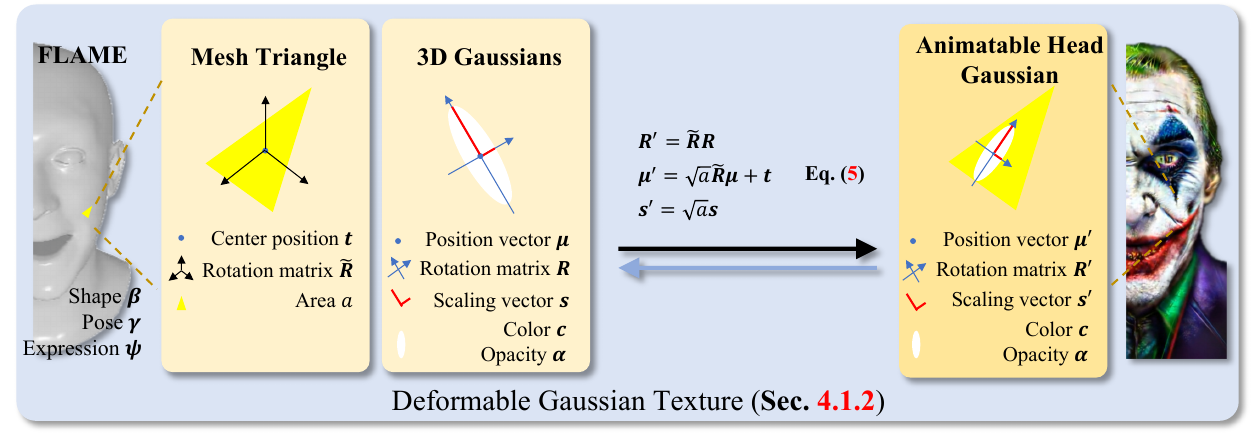}
    \caption{
    \textbf{The Details of Deformable Gaussian Texture.}
    Animatable head Gaussian uses the mesh triangle's center position, rotation matrix and area to translate, rotate and scale the corresponding rigged 3D Gaussians, resulting in a deformed 3D Gaussians.
    }
    \label{fig:overview-f-3dgs}
\end{figure}

\begin{figure}[t]
    \centering
    \includegraphics[width=1.0\linewidth]{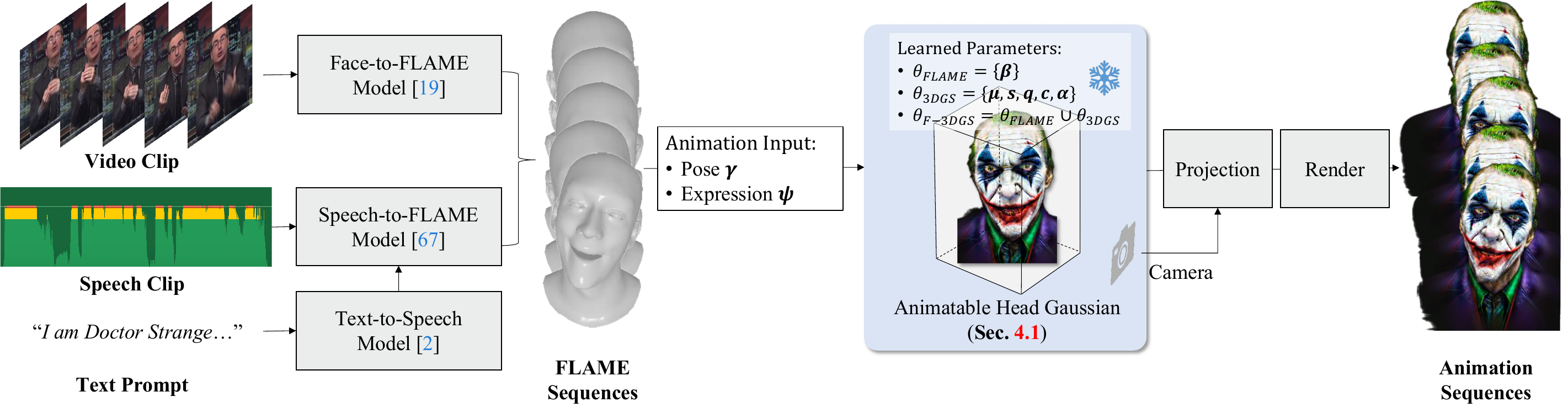}
    \caption{
    \textbf{The Pipeline of HeadStudio's Application.}
    The head avatar (fixed animatable head Gaussian) can be driven by video, speech, and text using FLAME pose and expression as control.
    }
    \label{fig:overview-inference}
\end{figure}

\noindent\textbf{0. Initialization. }
We evenly sample $K=10$ points per triangle from FLAME with the standard pose, and initialize the scaling via the square root of the mean distance of K-nearest neighbor points. 
The 3D Gaussians rigged with a large mesh triangle are initialized with a larger radius, compared to the ones rigged with a small mesh.
As a result, it initializes 3D Gaussians that can thoroughly cover the head model.
The further discussion of $K$ selection can be found in \cref{sec:supp-abl-init}.

\noindent\textbf{1. Random camera and animation sampling.}
At each iteration, the animation inputs, pose and expression are sampled from the FLAME sequences (pre-calculated based on the real-world talk show videos~\cite{yi2022generating}).
Meanwhile, a camera position is randomly sampled as described in Sec.~{\color{red}{4.3.3}}.

\noindent\textbf{2. Deform and render animatable head Gaussian.}
We detail the deformation process in \cref{fig:overview-f-3dgs}.
Given the pose and expression, FLAME with learnable shape is driven according to \cref{eq:flame}, deforming the mesh triangles.
Then, we utilize the mesh triangle's center position, rotation matrix and area to translate, rotate and scale the corresponding rigged 3D Gaussians (\cref{eq:f-3dgs}).
Following this, we render the deformed 3D Gaussians at a resolution of $1024 \times 1024$ based on the sampled camera pose.

\noindent\textbf{3. Optimization with animation-based text-to-3D distillation.}
Based on the FLAME model, we initially draw a facial landmark map in MediaPipe format as the diffusion condition. 
Then, we calculate the gradients of \cref{eq: f-sds} \wrt the animatable head Gaussian parameters, which force the rendering to satisfy the text prompt in any pose, expression, and camera view.

\noindent\textbf{4. Optimization with geometry regularization.}
We constrain the position and radius of 3D Gaussians \wrt the size of their rigged mesh triangle according to the \cref{eq:reg-naive}. 
Furthermore, an adaptive scaling factor is introduced in \cref{eq:reg} for modeling elements outside the space of FLAME.
The impact of the regularization is discussed in \cref{sec:supp-abl-reg}.

\subsection{The Pipeline of HeadStudio's Application}
We present the pipeline of HeadStudio's application in \cref{fig:overview-inference}.
Once optimized, the parameters of the avatar remain fixed.
Given a pose and expression, it can be deformed and rendered in a novel view.
Combined with advanced techniques, such as face-to-FLAME model~\cite{DECA:Siggraph2021}, speech-to-FLAME model~\cite{yi2022generating} and text-to-speech model~\cite{PlayHT}, the video, speech and text can be converted into FLAME animation inputs.
HeadStudio processes the input frame by frame and produces the animation sequences, which can then be merged into a video.
Consequently, HeadStudio can be driven by multi-modality and achieves real-world applications (as shown in supplementary videos).

\begin{figure}[t]
    \centering
    \includegraphics[width=1.0\linewidth]{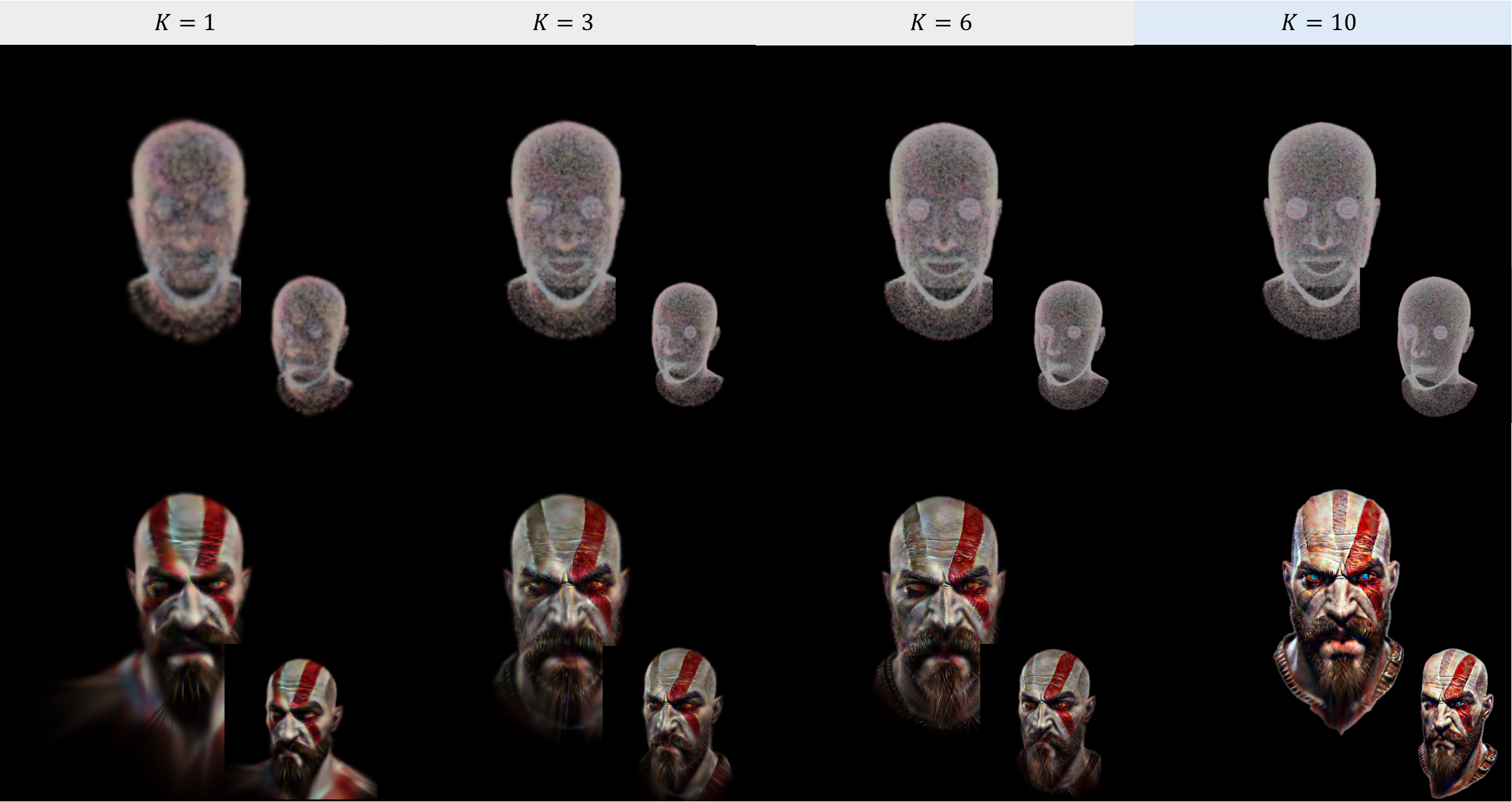}
    \caption{
    \textbf{Evaluation on $K$ in super-dense Gaussian initialization.}
    The cloning and splitting strategy can not handle the generation well.
    Increasing $K$ improves generation results with dense initialization.
    }
    \label{fig:supp-abl-k}
\end{figure}

\section{Additional Experiments}

\begin{figure}[t]
    \centering
    \includegraphics[width=0.8\linewidth]{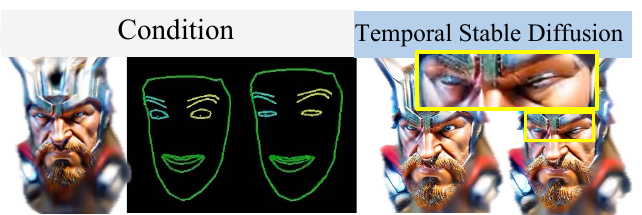}
    \caption{
    \textbf{Evaluation on temporal stable diffusion.}
    The temporal information is important to improve the temporal smoothness (skin wobbles) and animation quality (never blinking).
    }
    \label{fig:abl-temporal}
\end{figure}

\subsection{Temporal Stable Diffusion}
Temporal stable diffusion, such as AniPortrait~\cite{wei2024aniportrait}, introduces motion module into the denoising UNet.
As a result, it can generate a video clip with temporal consistency.
It inspires us to utilize a temporal stable diffusion to improve the temporal smoothness (skin wobbles) and animation quality (never blinking).
As shown in \cref{fig:abl-temporal}, the temporal information is indeed significant for generating smoother animations, and we will consider incorporating more temporal designs to enhance temporal supervision in the future.

\subsection{Additional Ablations}
\noindent\textbf{Evaluation on different $K$ in super-dense Gaussian initialization.}
\label{sec:supp-abl-init}
We discuss the impact of the hyperparameter $K$ in HeadStudio.
As shown in \cref{fig:supp-abl-k}, the proposed initialization is essential for generation.
In the default configuration ($K=1$), the animatable head Gaussian is unable to grow up through cloning and splitting~\cite{kerbl3Dgaussians}, leading to a poor appearance.
We attribute it to the sparse guidance provided by score distillation-based loss.
On the other hand, the density of 3D Gaussians is similar to the resolution of the image.
A denser 3D Gaussians will have a better representation ability.
Therefore, with the increase of $K$, the dense initialization results in better generation results.
However, a large $K$ will result in additional time and memory costs.
Therefore, we opt for $K=10$ as the default experimental setup.

\begin{figure}[t]
    \centering
    \includegraphics[width=1.0\linewidth]{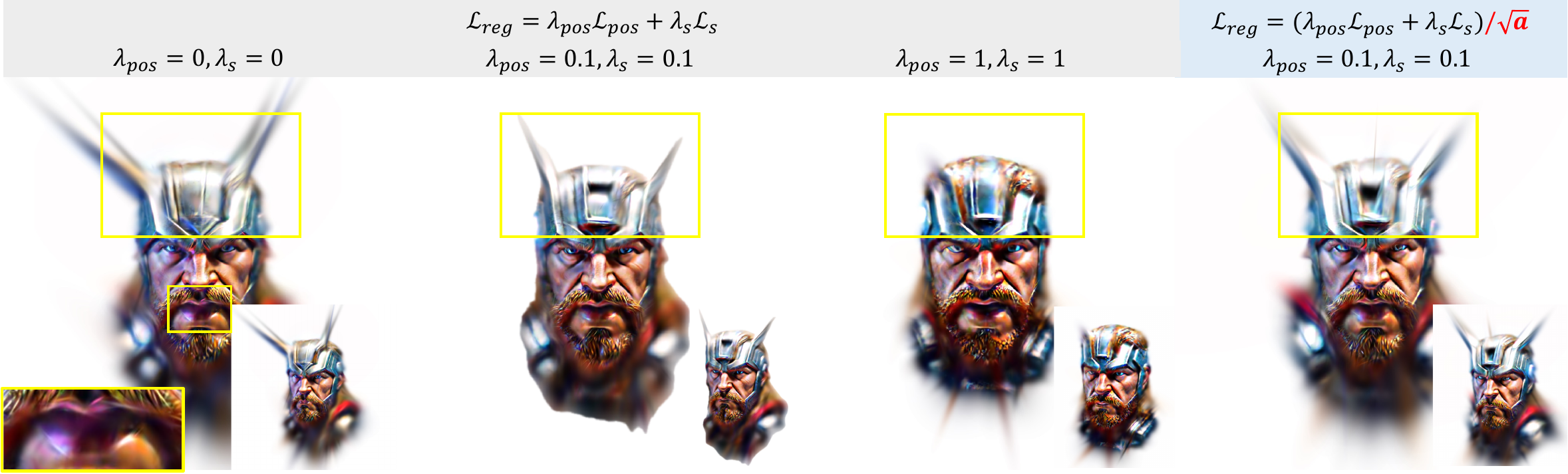}
    \caption{
    \textbf{Evaluation on adaptive geometry regularization.}
    Regularization is essential for semantic deformation.
    But the weight of regularization must find a good balance between alignment and representation.
    Including an adaptive scaling factor helps to combine semantic alignment and adequate representation well.
    }
    \label{fig:supp-abl-reg}
\end{figure}

\begin{figure}[t]
    \centering
    \includegraphics[width=1.0\linewidth]{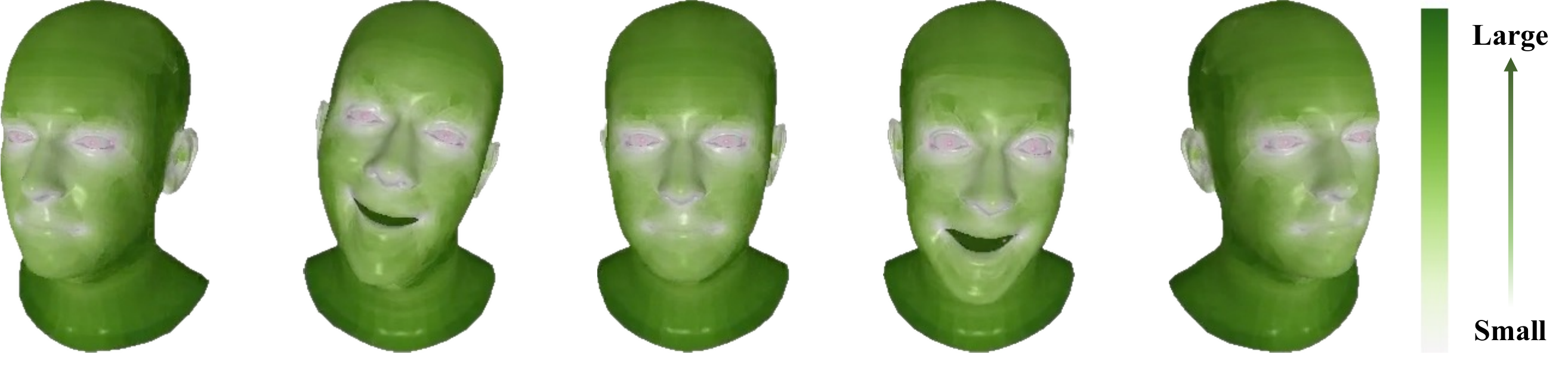}
    \caption{
    \textbf{More Visualization of Mesh Area.}
    We visualize the area of mesh triangle, where small mesh is white and large mesh is green.
    The mesh around the eyes, noise, mouth and ears is small, while the mesh on the jaw and above the head is relatively larger.
    }
    \label{fig:supp-abl-mesh-area}
\end{figure}

\noindent\textbf{Evaluation on Adaptive Geometry Regularization.}
\label{sec:supp-abl-reg}
First, we investigate geometry regularization and explore the impact of its weight in HeadStudio.
As depicted in \cref{fig:supp-abl-reg}, geometry regularization is crucial for semantic deformation.
In the absence of geometry regularization ($\lambda_{pos}=0, \lambda_{s}=0$), the 3D Gaussians fail to align semantically with FLAME, resulting in the problem of mouths sticking together (first column in \cref{fig:supp-abl-reg}).
On the other hand, the weight shows a trade-off between alignment and representation.
For instance, the Thor in the third column, generated with a large constraint weight, shows good alignment in the mouth but lacks representation (the helmet is missing).
Then, we analyze the proposed adaptive scaling factor.
We choose the area of the mesh triangle as an adaptive scaling factor (shown in \cref{fig:supp-abl-mesh-area}), which is small around the eyes and mouth, and large on jaw and over head.
With the help of the adaptive scaling factor, the generation demonstrates semantic alignment and adequate representation simultaneously (fourth column in \cref{fig:supp-abl-reg}).
It highlights the importance of the adaptive scaling factor in geometry regularization, which effectively balances the alignment and representation.

\begin{figure}[t]
    \centering
    \includegraphics[width=1.0\linewidth]{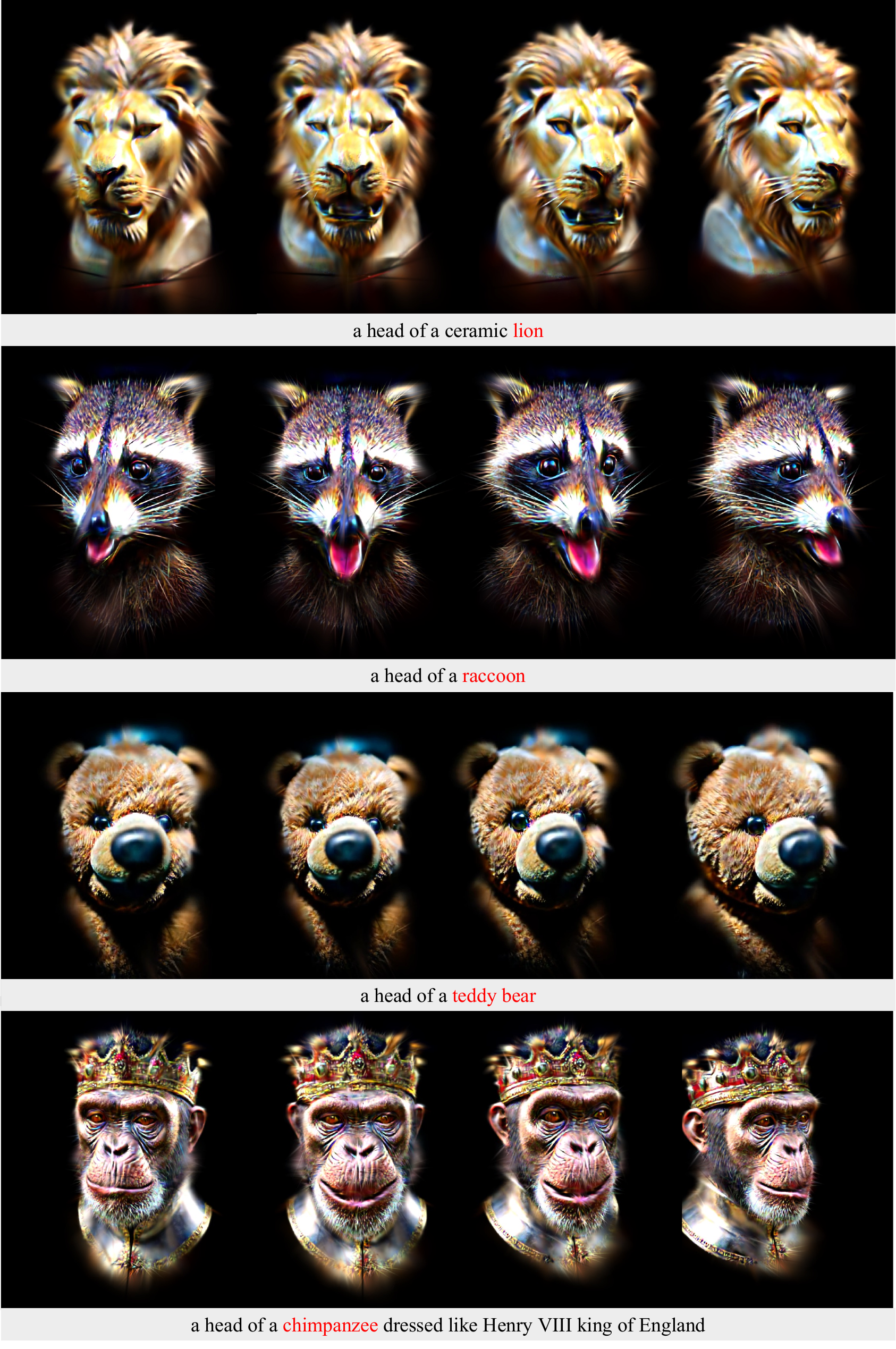}
    \caption{
    \textbf{Evaluation on Animal Character.}
    }
    \label{fig:supp-abl-animal}
\end{figure}

\begin{figure}[t]
    \centering
    \includegraphics[width=1.0\linewidth]{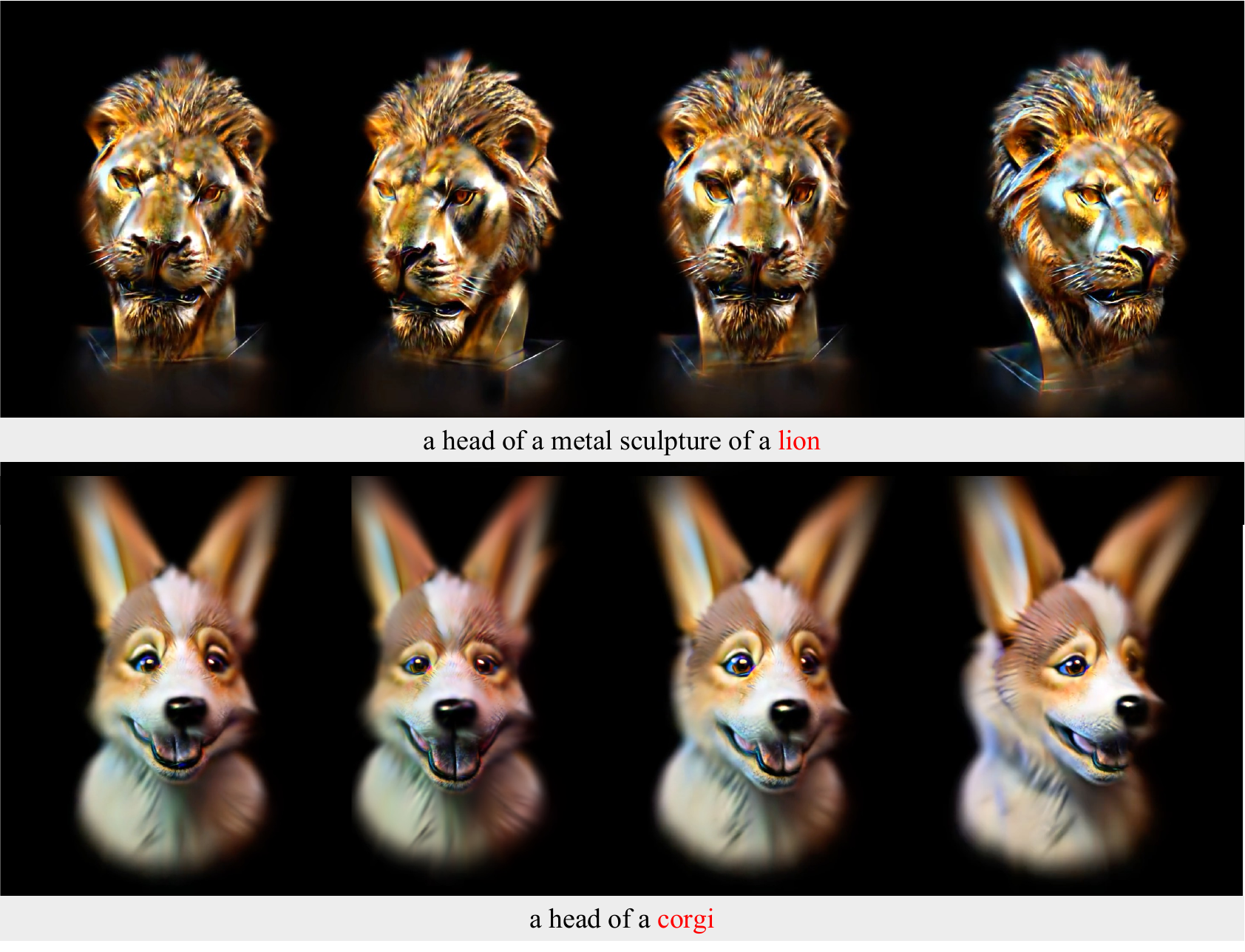}
    \caption{
    \textbf{Evaluation on Animal Character.}
    HeadStudio effectively generates animal characters, showing its versatility.
    }
    \label{fig:supp-abl-animal-2}
\end{figure}

\noindent\textbf{Evaluation on Animal Character.}
We evaluate the generalization of HeadStudio with various animal character prompts.
As shown in \cref{fig:supp-abl-animal-2} and \cref{fig:supp-abl-animal}, HeadStudio effectively generates animal characters, such as the lion, corgi, bear, raccoon and chimpanzee.
However, we believe that the human head prior model, FLAME~\cite{li2017flame}, could limit the animation quality.
In the future, replacing FLAME with an animal prior model like SMAL~\cite{zuffi2017smal} in HeadStudio could improve animal avatar generation.

\section{Limitations}
HeadStudio can create animatable head avatars from text for easier avatar production.
However, certain challenges need to be addressed before using avatars in applications.
For instance, it is important to develop a real-time driving and presentation system to integrate avatars into live broadcasts, which suited to 3DGS rendering pipeline.
For instance, to enable an avatar for live broadcasting, a real-time driving and presentation system suitable for 3DGS rendering should be developed.
Engineering issues such as complex workflows and audio-visual synthesis need to be carefully addressed.
Additionally, our method faces some limitations inherited from FLAME, particularly in representing teeth and hair.
Recent advancements in the teeth~\cite{qian2023gaussianavatars} and hair modeling~\cite{luo2024gaussianhair} could offer solutions to these limitations.

\end{document}